%% file: main.tex
\documentclass[10pt, letterpaper, logo, twocolumn, twoside]{noitom_robotics}

\usepackage{graphicx}
\usepackage{amsmath,amssymb}
\usepackage{booktabs}
\usepackage{hyperref}

\title{Towards Professional Tennis Styles for Humanoid Robots with Adaptive Motion Planning and Tracking}
\runningtitleEven{Stylized Humanoid Tennis with
Adaptive Motion Planning and Tracking}
\runningtitleOdd{AdaPT: \url{\website}}

\newcommand{\website}{https://humanoidtennis.github.io/AdaPT}
\renewcommand{\today}{August 2026}

\author[1,2,4,*,$\ddagger$]{Tao Huang}
\author[1,4,*]{Ruofei Liu}
\author[1]{Xuchen Tang}
\author[1]{Xinyin Zhang}
\author[2]{Junli Ren}
\author[2]{Huayi Wang}
\author[2]{Feiyu Jia}
\author[1]{Yukai Qi}
\author[2]{Kangning Yin}
\author[2]{Weishuai Zeng}
\author[4]{Lipeng Chen}
\author[3]{Xi Li}
\author[1]{Ting Wu}
\author[2]{Kailin Li}
\author[1]{Ruoli Dai}
\author[2,$\dagger$]{Jingbo Wang}
\author[1,$\dagger$,$\ddagger$]{Lei Han}
\author[2,$\dagger$]{Jiangmiao Pang}

\affil[1]{Noitom Robotics}
\affil[2]{Shanghai AI Laboratory}
\affil[3]{Dobot Robotics}
\affil[4]{Shanghai Jiao Tong University\affilbreak}
\affilnote{*}{Equal Contribution}
\affilnote{$\dagger$}{Equal Advising}
\affilnote{$\ddagger$}{Project Lead}

\input{preamble}
\input{math}

\begin{document}

\maketitle

\section*{\centering Abstract}
Humanoid robots have recently demonstrated promising capabilities in real-world ball sports. However, achieving professional motion styles while maintaining strong task performance remains challenging. In this work, we propose \textbf{\ours}, an \textbf{Ada}ptive motion \textbf{P}lanning and \textbf{T}racking framework that learns professional tennis serving and rally styles directly from broadcast videos. This hierarchical design is motivated by the key insight that the planner generates stylistic kinematic motions, while the tracker executes them with minimal interference with planning. Despite its effectiveness in simulation, a substantial sim-to-real gap emerges: tracking performance inevitably degrades on real robots, and this degradation is partially overlooked by autoregressive planning and further compounded by noisy perception. To address these issues, our adaptation mechanism improves tracking robustness by learning to track randomized execution speeds, while conditioning the planner on a learned motion-speed adapter to mitigate compounding errors. Real-world experiments on the Unitree G1 demonstrate the effectiveness of our adaptation mechanism in bridging the sim-to-real gap. We further deploy \ours policies on the full-size Dobot Atom humanoid robot ($\sim$1.7m) and demonstrate in-the-wild serving without motion capture. Beyond these results, our real-world experiments reveal both algorithmic and engineering insights for future humanoid ball-sports systems. Videos and code are available on our \href{https://humanoidtennis.github.io/AdaPT/}{project website}.

\input{sec/introduction}

\input{sec/related_work}
\input{sec/method}

\input{sec/experiments}

\section*{Acknowledgements}
We sincerely thank our teammates from the Intern Robotics team at Shanghai AI Laboratory, including Xiao Chen, Zirui Wang, Xiaojie Niu, Jiahe Chen, Weixiang Zhong, and Furui Xu, for their valuable suggestions on the early-stage preparation and algorithm design of this work. We are grateful to the Dobot engineering team, including Shiwen Liao, Yiliang Huang, Jiajun Wang, and Shuhai Jiang, for their support in designing the camera-based ball localization and robot spatial localization systems. In particular, we thank Tongbiao Cai for his patience and dedication in continuously refining the camera-based detection system and working alongside us through countless iterations and experiments. We also thank Wude Wang for his invaluable support in deploying our system on the Atom humanoid robot and resolving numerous hardware issues. We also thank the Motion Capture Team of Noitom Robotics and all the athletes who participated in our motion capture sessions for helping us rapidly collect high-quality rally and serve data, which was essential for our iterative development and evaluation. Finally, we dedicate this work to Roger Federer, Rafael Nadal, Novak Djokovic, and all the great tennis players whose perseverance, passion, and refusal to give up have continually inspired us to move forward.

\bibliography{example}

\input{sec/appendix}


\end{document}

%% file: preamble.tex
\usepackage{multirow}
\usepackage{multicol}
\usepackage{graphicx}
\usepackage{xspace}
\usepackage{xcolor}
\usepackage{caption}
\usepackage{wrapfig}
\usepackage{bbding}  
\usepackage{pifont}  

\usepackage{booktabs}
\usepackage{float}
\usepackage{amsmath}  
\usepackage{amssymb}  
\usepackage{bm}
\usepackage{dsfont}
\newcommand{\ours}[0]{AdaPT\xspace}

\usepackage{hyperref}
\usepackage[capitalise, nameinlink]{cleveref}
\crefname{table}{Tab.}{Tabs.}
\Crefname{table}{Tab.}{Tabs.}

\crefname{section}{Sec.}{Secs.}
\Crefname{section}{Sec.}{Secs.}

\newcommand{\paragraphbegin}[1]{\vspace{-0.02in}\noindent\textbf{#1}}

\usepackage{colortbl}
\definecolor{ourcolor}{HTML}{99e0eb}
\definecolor{ourblue}{HTML}{27a2c3}

\definecolor{tablecolor}{HTML}{ccf2f5} 

\definecolor{tablecolor2}{HTML}{ffcdb4}
\definecolor{citecolor}{HTML}{fe7b5b}
\definecolor{grey}{rgb}{0.9, 0.9, 0.9}
\usepackage{amssymb}

\usepackage{listings}
\definecolor{gred}{rgb}{0.859,0.267,0.216}
\definecolor{ggreen}{rgb}{0.059,0.616,0.345}

\definecolor{deepblue}{HTML}{27a2c3}

\definecolor{deepred}{HTML}{fe7b5b}

\usepackage[font=small,labelfont=bf]{caption}

\usepackage[font=footnotesize,labelfont=bf]{caption}

\definecolor{citecolor}{HTML}{faa700} 
\definecolor{lblue}{HTML}{016bde} 
\definecolor{ogreen}{HTML}{2E7D32}
\definecolor{bred}{HTML}{f41400}
\definecolor{newbrown}{HTML}{795548}
\definecolor{tennisred}{HTML}{d32a24}
\definecolor{tennisblue}{HTML}{0f6dbe}
\definecolor{tennisgreen}{HTML}{4E8810}

\hypersetup{
    colorlinks=true,
    linkcolor=tennisblue,
    filecolor=tennisred,      
    urlcolor=NoitomPurple,
    citecolor=tennisred,
}

\newcommand{\ourrow}{\rowcolor{gray!7}}

\newcommand{\ci}[1]{\tiny{\textcolor{gray}{$\pm #1$}}}

\usepackage{multicol}
\usepackage{multirow}
\usepackage{colortbl}
\usepackage{booktabs}   
\usepackage{bbding} 
\usepackage{graphicx}
 
\usepackage{enumitem}

\usepackage{bbding}

\usepackage{pifont}
\usepackage{xfrac}
\usepackage{tikz}

\usepackage[para,online,flushleft]{threeparttable}
\newcommand{\thickcdot}{
  \mathbin{\raisebox{0.15ex}{\scalebox{0.8}{$\bullet$}}}%
}

\newcommand{\nadalfull}[0]{\textcolor{tennisred}{Rafael Nadal}}

\newcommand{\federerfull}[0]{\textcolor{NoitomPurple}{Roger Federer}}

\newcommand{\djokovicfull}[0]{\textcolor{tennisblue}{Novak Djokovic}}

\usepackage{tcolorbox}
\newcommand{\takeaway}[1]{ {
    \begin{tcolorbox}[colback=gray!20, colframe=gray!60,left=1.2mm,right=1.2mm,top=1.3mm,bottom=1.3mm,boxsep=0mm]
\textit{Key trade-off}: #1 \end{tcolorbox}}}  

%% file: math.tex
\usepackage{amsfonts}
\usepackage{bm}
\usepackage{amsmath}
\usepackage{mathtools}
\usepackage{multirow}
\usepackage{booktabs}
\usepackage{caption}
\usepackage{wrapfig,lipsum,booktabs}
\usepackage{caption}
\usepackage{bbm}
\usepackage{multirow}
\usepackage{pifont}
\usepackage{tablefootnote}
\usepackage{threeparttable} 

\newcommand{\bs}[1]{\boldsymbol{#1}}

\definecolor{goldenyellow}{rgb}{0.99, 0.76, 0.0}
\usepackage{cleveref}
\usepackage{float}

%% file: sec/introduction.tex
\vspace{-0.03in}
\section{Introduction}
Humanoid robots have recently demonstrated promising capabilities in dynamic ball sports, including tennis~\cite{zhang2026learning}, table tennis~\cite{Hitter2025,hu2025towards,ren2026smash}, and badminton~\cite{hwb,chen2026learning}. Existing work mainly focuses on task performance (e.g., hitting and ball return) while paying less attention to professional styles. Such styles have been proven not merely aesthetic, but reflect structured whole-body coordination that is critical for efficient force generation and rapid recovery in human sports biomechanics~\cite{elliott2006biomechanics,kibler2004kinetic}. For humanoid robots, we envision that these styles can be leveraged to improve motion efficiency and task performance, embracing their temporally limited actuation and energy constraints.

In this work, we study tennis, a challenging ball sport involving diverse strokes, high-speed interaction, and full-body coordination. We first construct a lightweight pipeline to extract professional-player motions from broadcast videos and annotate stroke type, spin type, and serving toss timing. To preserve motion style while ensuring task robustness, we adopt a decoupled planning-and-tracking framework~\cite{ling2020character,zhang2023learning,huang2025towards,xu2025parc}, separating stylistic high-level motion generation from low-level execution. For rallying, we built on Vid2Player3D~\cite{zhang2023learning} and train an MVAE-based motion generator~\cite{ling2020character,zhang2023learning} and a planner that infers motion latents and tracking speed from predicted future ball trajectories. For serving, we build on AdaMimic~\cite{huang2025towards} and train a residual tracker to balance task performance and stylistic fidelity.
While this decoupled design is effective in simulation, a substantial sim-to-real gap emerges when deploying on real humanoid systems. In particular, tracking performance inevitably degrades due to imperfect actuator modeling in simulation~\cite{hwangbo2019learning,he2025asap}, and such errors are further amplified by the autoregressive nature of motion planning and perception noise of hardware, leading to compounding deviations over long-horizon execution~\cite{parkourpt}.

To address these issues, we introduce \textbf{\ours} (\textbf{Ada}ptive Motion \textbf{P}lanning and \textbf{T}racking), a framework that jointly models motion planning and tracking with explicit speed adaptation. It improves tracking robustness by exposing the policy to randomized motion execution speeds during training. Meanwhile, the high-level planner learns a motion-speed representation, enabling it to adapt planning to varying tracking capabilities and mitigate error accumulation in real-world deployment. 

We validate \ours on both the Unitree G1 and the full-size Dobot Atom humanoid robots. Using broadcast video data, \ours captures and reproduces the distinctive playing styles of three professional players—Rafael Nadal, Roger Federer, and Novak Djokovic—for both rallying and serving. We further show that \ours generalizes to motion-capture (MoCap) data, demonstrating its applicability across diverse motion sources. Beyond controlled laboratory settings, we deploy \ours in the wild, achieving successful serving with a YOLO-based~\cite{redmon2016you} camera perception system for ball localization and HTC VIVE Ultimate Trackers~\cite{HTC2024VIVEUltimateTracker} for robot localization. Together, these results demonstrate the robustness, versatility, and practical deployment potential of \ours for future in-the-wild humanoid ball sports.

We overview the real-world performance in \cref{fig:teaser} and summarize our core contributions as follows: 

\begin{itemize}[leftmargin=4mm, itemsep=0.25pt, topsep=0pt]
    \vspace{-0.03in}
    \item \textbf{A unified adaptive motion planning and tracking framework} that mitigates the sim-to-real gap by improving robustness to tracking errors and enabling the planner to adapt to the tracker's capabilities.
    \vspace{-0.02in}
    \item \textbf{Professional humanoid tennis across motion sources and embodiments} is demonstrated through rallying and serving from both video and motion-capture data, on both the Unitree G1 and full-size Dobot Atom robots.
    
    \vspace{-0.02in}
    \item \textbf{Real-world deployment beyond motion capture} is further demonstrated via successful in-the-wild serving.
    
    \vspace{-0.02in}
    \item \textbf{Algorithmic and engineering insights} from real-world experiments that provide practical design guidance for future humanoid ball-sports systems.
\vspace{-0.03in}
\end{itemize}

%% file: sec/related_work.tex
\section{Related Work}
\paragraph{Robotic ball sports.} Robotic systems have demonstrated impressive capabilities in a variety of ball sports~\cite{huang2023creating,tirumala2024learning,riedmiller2009reinforcement,mori2018high,MaCFH25,wang2025integrating,buchler2022learning,dambrosio2025achieving,zaidi2023athletic}. Recent humanoid systems have achieved promising performance in football~\cite{haarnoja2024learning,ren2025humanoid,wang2025learning}, basketball~\cite{wang2026humanx}, badminton~\cite{hwb,chen2026learning}, table tennis~\cite{Hitter2025,ren2026smash,hu2025towards}, and tennis~\cite{zhang2026learning}, primarily focusing on task success such as hitting and ball returning. In contrast, we investigate how humanoid robots can acquire professional athletic styles while maintaining robust task execution, a setting that introduces a substantially larger sim-to-real challenge due to the increased requirements on timing, coordination, and force generation. We further study stylized humanoid serving, a whole-body skill that remains largely unexplored in real-world humanoid ball sports.

\paragraphbegin{Athletic humanoid motion learning.} Recent works have explored learning athletic humanoid skills from motion data. AdaMimic~\cite{huang2025towards} demonstrates adaptable motion tracking from a single reference motion, but does not incorporate perception and has not been applied to dynamic ball sports. We extend its adaptation mechanism to stylized tennis serving, improving robustness to ball toss variations while reducing policy optimization difficulty. For ball sports, PULSE~\cite{luo2024universal} introduces a continuous motion prior that has been successfully adopted in various downstream tasks, including the ball sports benchmarks in SMPLOlympics~\cite{luo2024smplolympics}. LATENT~\cite{zhang2026learning} further extends this paradigm to real humanoid tennis. While these methods achieve strong task performance, the motion prior is tightly coupled with environment dynamics, making it difficult to preserve professional motion styles from references. Vid2Player3D~\cite{zhang2023learning} addresses this issue through decoupled motion planning~\cite{ling2020character} and tracking, where planning focuses solely on generating kinematic motions. However, we find that this decoupled formulation suffers from a substantial sim-to-real gap during real-world deployment, as tracking degradation, autoregressive planning errors, and perception noise jointly lead to compounding execution errors. Building on Vid2Player3D, we introduce adaptive planning and tracking mechanisms to mitigate these issues, enabling robust deployment on real humanoid robots. Furthermore, we provide empirical insights into the trade-offs between the above approaches for real-world humanoid tennis.

%% file: sec/method.tex
\section{AdaPT: \underline{Ada}ptive Motion \underline{P}lanning and 
\underline{T}racking for Humanoid Tennis}

\begin{figure*}[tbp]
    \centering
    \includegraphics[width=1\linewidth]{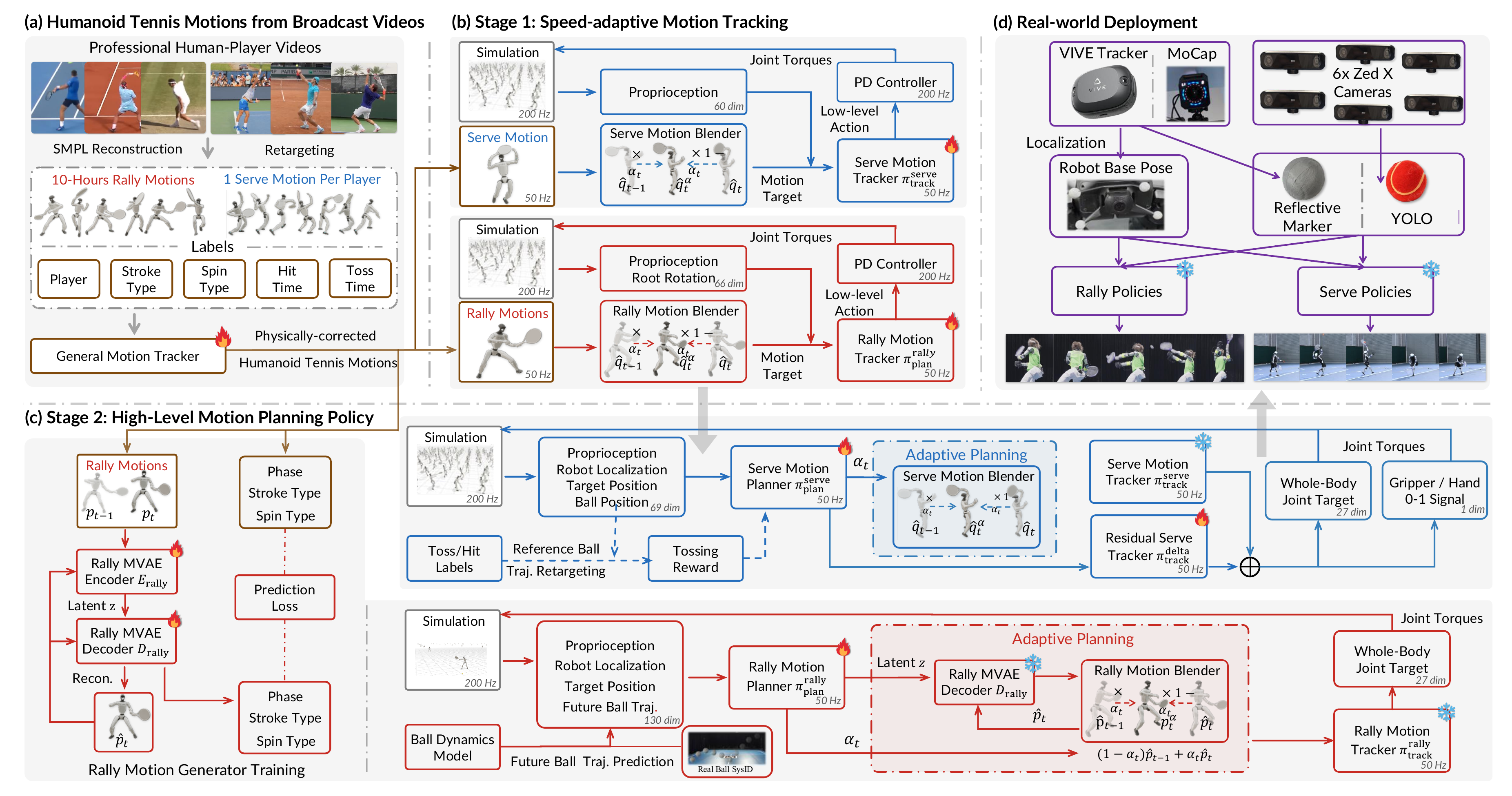}
    \vspace{-0.25in}
    \caption{
    \ours learns stylized humanoid tennis skills through the pipeline: (a) broadcast videos are reconstructed, retargeted, and physically corrected to obtain tennis motions;
    (b) speed-adaptive policies learn to track serve and rally motions; (c) high-level planners select or generate style-consistent motions for serve and rally; (d) our robot localization or ball detection modules enable sim-to-real deployment, especially for serving without MoCap.
    }
    \label{fig:method}
    \vspace{-0.15in}
\end{figure*}

Our goal is to learn professional humanoid tennis skills through the decoupled motion planning and tracking framework, where the motion planning is performed independently of environment dynamics and focuses on generating kinematic motions to ensure stylistic imitation (\cref{fig:method}). Intuitively, rally and serve exhibit fundamentally different characteristics, motivating distinct method designs.

\begin{wrapfigure}{r}{0.42\linewidth}
    \centering
    \vspace{-0.15in}
    \includegraphics[width=1\linewidth]{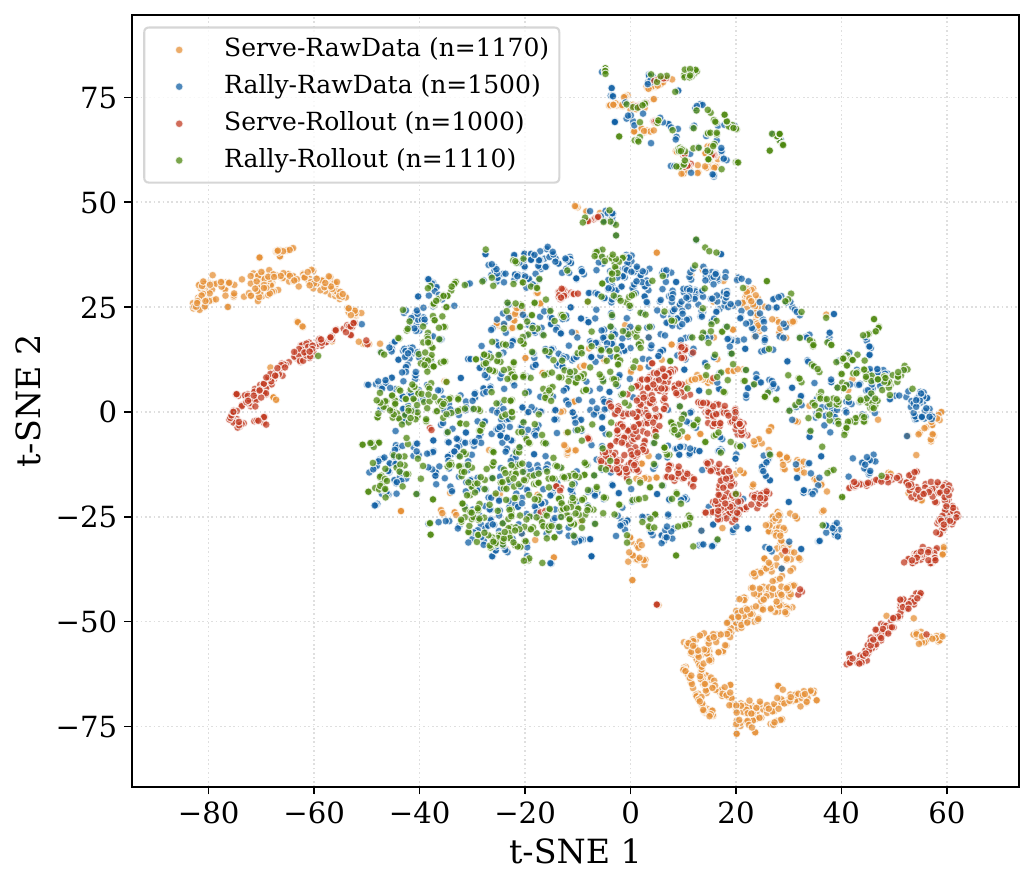}
    \vspace{-0.28in}
    \caption{t-SNE visualization of motions and tracker rollouts.}
    \vspace{-0.2in}
    \label{fig:tsne}
\end{wrapfigure}

\paragraphbegin{Characteristics of rally and serve.} For rally skills, the robot must react to unknown incoming ball trajectories and generate diverse motions that can cover large return regions while supporting different stroke types. In contrast, serving is a self-initiated process from ball toss to racket contact, which is largely independent of external environment dynamics and exhibits much lower motion diversity, as verified by the motion distribution shown in~\cref{fig:tsne}.  
From a reinforcement learning (RL) perspective, they correspond to self-play with an unknown opponent policy and standard single-agent RL problems, respectively.

\paragraphbegin{Methodology choices.}
Based on these observations, we build the rally method upon Vid2Player3D~\cite{zhang2023learning} and learn an MVAE-based motion generator~\cite{ling2020character} to provide the planner with diverse motion generation capabilities. For serve, we instead build upon AdaMimic~\cite{huang2025towards}, where we train a residual tracker to better balance tasks and styles.

\paragraphbegin{Overview of \ours.}
With the curated tennis motion dataset (\cref{subsec:motion-data}), we train motion trackers using randomized speeds to enable speed-adaptive tracking (\cref{subsec:traker}). We then train planners with an extra planning dimension on tracking speed for improved task performance while preserving styles (\cref{subsec:planner}), followed by the details of (\cref{subsec:training-details}) and real-world deployment (\cref{subsec:deployment}).

\subsection{Motion Data from Broadcast Videos}
\label{subsec:motion-data}

\paragraph{Collection.}
To learn professional tennis styles, we collect short tennis clips from publicly available broadcast videos of Rafael Nadal, Roger Federer, and Novak Djokovic, given their distinctive playing styles and the availability of sufficient public videos. Each clip is approximately 2 seconds long and contains a complete rally motion and associated footwork. In addition, we collect high-fidelity motion-capture (MoCap) data from professional human tennis athletes to provide additional styles, demonstrating the compatibility of our framework with diverse motion data sources. We list the statistics in~\cref{tab:data_dist}.

\begin{table}[h]
\centering

\vspace{-0.1in}
\caption{Statistics of motion data from broadcast videos of three players.}
\label{tab:data_dist}

\setlength{\tabcolsep}{5pt}
\renewcommand{\arraystretch}{1}

\vspace{-0.1in}
\resizebox{1\linewidth}{!}{
\begin{tabular}{l|cc|cc|cc}
\toprule

\multirow{2}{*}{Player} &
\multirow{2}{*}{\shortstack{Total \\ Time}} &
\multirow{2}{*}{\shortstack{\# of \\ Clips}} &

\multicolumn{2}{c|}{Stroke Ratio} &
\multicolumn{2}{c}{Spin Ratio} \\ [-0.3ex]
\cmidrule(lr){4-5}
\cmidrule(lr){6-7}
& & &
Forehand & Backhand &
Flat \& Top & Slice \\[-0.3ex]

\midrule

\nadalfull     & 267min & 5664 & 54\% & 46\% & 89\% & 11\% \\
\federerfull   & 179min & 5264 & 48\% & 52\% & 84\% & 16\% \\
\djokovicfull  & 189min & 5288 & 49\% & 51\% & 93\% & 7\% \\

Mr. Black (MoCap) & 219min & 2987 & 57\% & 43\% & 76\% & 24\% \\
\bottomrule
\end{tabular}
}

\vspace{-4mm}
\end{table}
\paragraphbegin{Processing.} We reconstruct SMPL~\cite{SMPL} motions from the videos using GVHMR~\cite{Shen2024WorldGroundedHM} and retarget them to the humanoid robot with GMR~\cite{joao2025gmr}. Since GVHMR often produces inaccurate wrist estimates due to viewpoint occlusions, we refine racket-related wrist joints according to player-specific stroke styles following Vid2Player3D. We further apply randomized wrist perturbations during retargeting to diversify racket swing directions and reduce model biases. This is only applied to the rally data.

\paragraphbegin{Labeling.}
Each motion clip is annotated with player identity, stroke type, spin type, and ball contact timing. The stroke labels include forehand, backhand, and serve, while the spin labels distinguish between topspin and slice. These labels provide additional motion semantics for policy learning. For serving motions, we additionally annotate the ball release timing to estimate the toss trajectory.

\paragraphbegin{Correction.}
Since the retargeted robot motions may violate physical constraints and are not directly suitable for tracking, we train a general motion tracker on the retargeted dataset to refine them into physically plausible motion clips, resulting in the final motions $\mathcal{D}_{\mathrm{rally}}$ and  $\mathcal{D}_{\mathrm{serve}}$.

\subsection{Speed-adaptive Motion Tracking}\label{subsec:traker}

\paragraph{Problem formulation.}
We formulate motion tracking as a Markov decision process~\cite{puterman2014markov}. At timestep $t$, the observation is defined as
$\bs{o}^{\mathrm{track}}_t =
[\hat{\bs{q}}_{t}, \bs{q}_{t}, \dot{\bs{\theta}}_t, \dot{\bs{\omega}}_t],$
where $\hat{\bs{q}}_{t}$ and $\bs{q}_{t}$ denote the reference and current robot configurations, respectively, each comprising the root orientation and joint positions, following~\cite{Liao2025BeyondMimicFM, wang2026omnixtreme}. $\dot{\bs{\theta}}_t$ and $\dot{\bs{\omega}}_t$ denote the robot's joint velocities and base angular velocity, respectively. We find that incorporating root orientation is particularly important for learning rally styles, as it helps capture torso rotation during stroke execution. In contrast, this information is less critical for the relatively structured serving motions and is therefore omitted from the serving tracker. The policy outputs a tracking action
$
\bs{a}^{\mathrm{track}}_t = \pi_{\mathrm{track}}(\bs{o}^{\mathrm{track}}_t),
$
conditioned on the current observation. The action $\bs{a}^{\mathrm{track}}_t$ is executed as the target command of a PD controller~\cite{wang2026omnixtreme} during each control period. The agent receives a reward $r_t$ to minimize global and local tracking errors.

\paragraphbegin{Objective.}
Our speed adaptation builds on the idea introduced in AdaMimic~\cite{huang2025towards}. While AdaMimic does not pre-train the tracker to handle varying motion speeds, we find such adaptability to be important for real-robot performance. To this end, we randomly vary the speed of the reference motion by blending the current and previous reference states:
$
\hat{\bs{q}}_{t}^{\alpha}
=
(1-\alpha)\hat{\bs{q}}_{t-1}
+
\alpha\hat{\bs{q}}_{t},
$
where the blend coefficient $\alpha$ is uniformly sampled from $[\alpha_{\mathrm{min}},\alpha_{\mathrm{max}}]$.The blending interpolates or extrapolates the reference motion, effectively slowing down or speeding up its temporal progression. This exposes the policy to varying execution speeds during training and improves its robustness at deployment. The training objective is defined as
$ \mathbb{E}_{\pi_{\mathrm{track}}}\left[ \sum_{t=0} \gamma^t r_t \;|\; \hat{\bs{q}}_{t}^\alpha, \alpha \sim \mathrm{Unif.(\alpha_{\mathrm{min}}, \alpha_{\mathrm{max}})} \right], $
We train trackers $\pi_{\mathrm{track}}^{\mathrm{rally}}$ and $\pi_{\mathrm{track}}^{\mathrm{serve}}$ for rally and serve.

\subsection{Motion Planning with Speed Adaptation}\label{subsec:planner}

\subsubsection{Learning Rally with Motion Generator}

\paragraph{Motion generator.}
To model the diverse motions required for rally skills, we follow Vid2Player3D~\cite{zhang2023learning} and train an MVAE-based motion generator~\cite{ling2020character} on the corrected motion dataset $\mathcal{D}_{\mathrm{rally}}$. Please refer to their paper for more implementation details. In addition, we augment the training objective with auxiliary prediction losses on stroke type and spin type, which improves controllability through downstream reward design. After training, we obtain an MVAE decoder $D_{\mathrm{rally}}$, which takes as input a latent variable $\bs{z}_t$ together with the autoregressive state $\hat{\bs{p}}_{t-1}^{\mathrm{mvae}}$, and predicts the next motion state:
$\hat{\bs{p}}_{t}^{\mathrm{mvae}} = D_{\mathrm{rally}}(\bs{z}_t, \hat{\bs{p}}_{t-1}^{\mathrm{mvae}}).$

\paragraphbegin{High-level planning policy.}
On top of the low-level speed-adaptive tracker, we train a high-level planning policy $\pi^{\mathrm{rally}}_{\mathrm{plan}}$ that predicts both motion generation commands and tracking speed adaptation signals. Specifically, compared with the tracker, the planning policy additionally receives observations of the robot pose and future ball trajectory estimation and outputs a latent motion command for the MVAE decoder together with a speed adaptation variable $(\bs{z}_t,\alpha_t)=\pi_{\mathrm{plan}}^{\mathrm{rally}}(\bs{o}_t^{\mathrm{rally}})$. The speed adaptation signal $\alpha_t$ allows the planner to dynamically adjust motion execution speed according to the incoming ball trajectory. The reference motion for tracker at timestep $t$ is $
    \hat{\bs{q}}_t^{\mathrm{rally}} = f((1-\alpha_t)\cdot \hat{\bs{p}}_{t-1}^{\mathrm{mvae}} + \alpha_t \cdot \hat{\bs{p}}_{t}^{\mathrm{mvae}} ),$
where $f$ converts the MVAE motion representation into the tracker reference state. The weighted MVAE state is then used to condition the next stage of generation.

\vspace{-0.05in}
\subsubsection{Learning Serving with Residual Tracker}

\paragraph{High-level planning policy.}
Unlike rally, serving motions exhibit significantly lower motion diversity and are largely self-driven. Therefore, instead of learning an additional motion generator, we directly track motions from the serving dataset $\mathcal{D}_{\mathrm{serve}}$. On top of the low-level tracker, we train a high-level planning policy $\pi_{\mathrm{plan}}^{\mathrm{serve}}$ to adapt motion execution speed during serving. Specifically, the planner receives observations of the robot state, ball state, and its last action, and predicts a speed adaptation variable
$
\alpha_t = \pi_{\mathrm{plan}}^{\mathrm{serve}}(\bs{o}_t^{\mathrm{serve}}).
$
The resulting tracking reference $
\hat{\bs{q}}_t^{\mathrm{serve}} =
(1-\alpha_t)\hat{\bs{q}}_{t-1}
+
\alpha_t \hat{\bs{q}}_{t}
$ is then constructed by interpolating adjacent serving references.

\paragraphbegin{Residual motion tracker.}
Directly tracking serving motions alone is insufficient, since real-world toss trajectories inevitably introduce variations that cannot be fully handled by fixed reference tracking. To address this issue, we build upon AdaMimic~\cite{huang2025towards} and train a residual tracker that predicts an additional correction term on top of the reference-tracking action:
$
\bs{a}^{\mathrm{serve}}_t=
\bs{a}^{\mathrm{track}}_t
+
\Delta \bs{a}_t,$
where $\Delta \bs{a}_t = \pi^{\mathrm{delta}}_{\mathrm{track}}(\bs{o}^{\mathrm{serve}}_t)$ is a residual action that enables adaptive racket control under toss variations while preserving the underlying serving style.

\paragraphbegin{Keyframing.}
We observe that, when optimizing primarily for the striking reward, the policy may converge to a local optimum that neglects the backswing, resulting in less expressive serving motions. A similar issue was observed in AdaMimic~\cite{huang2025towards} for jumping tasks, where task rewards alone can favor solutions that achieve the objective while deviating from the desired motion style. To address this issue, we designate the frame with the deepest backswing as a keyframe and introduce an additional sparse local tracking reward around this phase. Specifically, we impose local tracking objectives on the racket and wrist poses, with a reward weight 50$\times$ larger than that of the standard tracking reward. This keyframe-guided supervision encourages the policy to preserve the characteristic backswing while retaining flexibility in the remaining motion phases.

\paragraphbegin{Tossing guidance.} 
We model the tennis toss trajectory as a parabola~\cite{tossingbot/trob/ZengSLRF20}, where the ball release and striking positions define the start and end points of the trajectory. Assuming gravity-only ballistic motion during the ascending phase, we estimate the expected toss velocities from the release-to-strike duration in the data. The dense tossing reward is computed as the deviation between the actual ball trajectory and the predicted parabolic trajectory over time.

\begin{wrapfigure}{r}{0.4\linewidth}
    \centering
    \vspace{-0.2in}
    \includegraphics[width=1\linewidth]{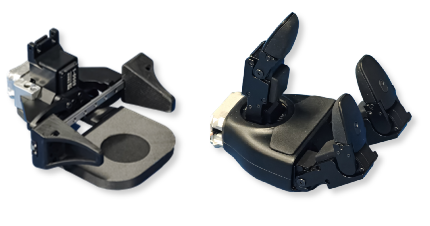}
    \vspace{-0.15in}
    \caption{Serve hardware.}
    \vspace{-0.16in}
    \label{fig:gripper}
\end{wrapfigure}
\paragraphbegin{Ball-tossing control.} 
During the tossing, the end-effector undergoes two phases: grasping and release. We set it closed before the tossing time and open afterward, allowing our serving system to accommodate different end-effector hardware (examples in \cref{fig:gripper}). To overcome real-world end-effector control latency and simulation inaccuracies in toss collisions, we randomize the release timing and randomly perturb ball velocity after release during training.

\begin{table*}[tbp]
    \centering
    \setlength{\tabcolsep}{1.5pt}
    \renewcommand\arraystretch{1.1}
     \caption{\textbf{Simulation results.} 
     Evaluation of all candidate methods across different metrics for rally and serve tasks by learning the motion of three master-level players. 
     $\uparrow$ denotes higher is better, while $\downarrow$ denotes the opposite. Overall, AdaPT consistently achieves strong motion fidelity while maintaining high task performance, demonstrating an effective balance between stylistic imitation and task performance across both rallying and serving.
     }
     \vspace{-0.1in}
    \captionsetup{justification=centering, singlelinecheck=false}
    \resizebox{\textwidth}{!}{
    \begin{threeparttable}
  \begin{tabular}{l || ccccc | ccccc | ccccc}
    \toprule
    \multirow{2}{*}{{\textbf{Stroke}}} &
      \multicolumn{5}{c|}{\textbf{\nadalfull}} &
      \multicolumn{5}{c|}{\textbf{\federerfull}} &
      \multicolumn{5}{c}{\textbf{\djokovicfull}}  \\ [-0.5ex]
    & Succ. $\uparrow$ & $E_{\mathrm{bo}} \downarrow$ & $E_\mathrm{FID} \downarrow$ & $E_\mathrm{pow} \downarrow$ & $E_{\mathrm{acc}} \downarrow$ 
    & Succ. $\uparrow$ & $E_{\mathrm{bo}} \downarrow$ & $E_\mathrm{FID} \downarrow$ & $E_\mathrm{pow} \downarrow$ & $E_{\mathrm{acc}} \downarrow$  
    & Succ. $\uparrow$ & $E_{\mathrm{bo}} \downarrow$ & $E_\mathrm{FID} \downarrow$ & $E_\mathrm{pow} \downarrow$ & $E_{\mathrm{acc}} \downarrow$  
    \\
    \hline
    RL-Scratch~\cite{schulman2017proximal}
        & \textbf{97.4} & \textbf{1.6}\ci{0.6} & 297.3\ci{16.3} & 92.7\ci{20.0} & 123.1\ci{8.9} &
        \textbf{98.8} & 2.5\ci{1.8} & 299.0\ci{29.8} & 130.8\ci{27.9} & 145.0\ci{8.3} &
        \textbf{98.8} & \textbf{2.5}\ci{1.8} & 307.7\ci{18.4} & 130.8\ci{27.9} & 145.0\ci{8.3}
        \\
    AMP~\citep{peng2021amp}
        & 98.1 & 2.7\ci{0.9} & 172.1\ci{8.0} & 118.1\ci{39.3} & 126.9\ci{12.5}
        & 65.1 & 2.1\ci{1.2} & 116.0\ci{32.3} & 112.7\ci{62.4} & 122.5\ci{16.7}
        & 21.6 & 2.2\ci{1.4} & 66.9\ci{39.7} & 185.6\ci{92.2} & 155.2\ci{57.7}
        \\
    PULSE~\citep{luo2024universal}
        & 86.2 & 2.5\ci{1.4} & 29.7\ci{3.5} & 92.2\ci{31.1} & 107.1\ci{11.4}
        & 92.2 & 2.5\ci{1.1} & 17.9\ci{1.5} & 94.1\ci{26.7} & 101.0\ci{9.5}
        & 86.1 & 3.4\ci{1.7} & 32.9\ci{6.3} & 78.6\ci{29.6} & 92.8\ci{7.5}
        \\
    NCP~\citep{zhu2023neural}
        & 85.6 & 4.8\ci{2.3} & 9.2\ci{3.5} & \textbf{33.0}\ci{13.2} & 55.3\ci{5.6}
        & 89.1 & 3.6\ci{1.5} & 6.8\ci{4.7} & \textbf{24.2}\ci{15.3} & 48.6\ci{5.2}
        & 69.0 & 4.9\ci{2.3} & 24.0\ci{2.8} & 22.8\ci{6.2} & 56.6\ci{4.6}
        \\  
    Vid2Player3D~\citep{zhang2023learning}
        & 82.6 & 4.8\ci{2.0} & \textbf{5.1}\ci{5.2} & 15.1\ci{5.6} & \textbf{41.1}\ci{5.6}
        & 81.5 & 2.9\ci{1.2} & 13.4\ci{3.2} & 18.5\ci{8.3} & \textbf{41.9}\ci{7.9}
        & 83.2 & 3.3\ci{1.9} & \textbf{6.3}\ci{3.2} & 24.0\ci{9.7} & \textbf{40.6}\ci{7.3}
        \\
    \ourrow\textbf{\ours (ours)} 
        & 91.5 & 2.9\ci{1.4} & 5.4\ci{2.8} & 16.3\ci{4.3} & 43.3\ci{5.0}
        & 96.3 & \textbf{2.2}\ci{1.1} & \textbf{6.1}\ci{1.3} & 25.2\ci{6.7} & 47.7\ci{4.1}
        & 92.3 & 2.8\ci{1.7} & \textbf{6.3}\ci{4.0} & \textbf{22.4}\ci{7.7} & 41.4\ci{6.4}
        \\
    \hline
    \hline
    \multirow{2}{*}{{\textbf{Serve}}} &
      \multicolumn{5}{c|}{\textbf{\nadalfull}} &
      \multicolumn{5}{c|}{\textbf{\federerfull}} &
      \multicolumn{5}{c}{\textbf{\djokovicfull}} \\
   & Succ. $\uparrow$ & $E_{\mathrm{bounce}} \downarrow$ & $E^{\mathrm{global}}_\mathrm{mpkpe} \downarrow$ & $E^{\mathrm{local}}_\mathrm{mpkpe} \downarrow$ & $E_{\mathrm{acc}} \downarrow$ 
   & Succ. $\uparrow$ & $E_{\mathrm{bounce}} \downarrow$ & $E^{\mathrm{global}}_\mathrm{mpkpe} \downarrow$ & $E^{\mathrm{local}}_\mathrm{mpkpe} \downarrow$ & $E_{\mathrm{acc}} \downarrow$ 
   & Succ. $\uparrow$ & $E_{\mathrm{bounce}} \downarrow$ & $E^{\mathrm{global}}_\mathrm{mpkpe} \downarrow$ & $E^{\mathrm{local}}_\mathrm{mpkpe} \downarrow$ & $E_{\mathrm{acc}} \downarrow$ 
        \\
    \hline
    RL-Scratch~\cite{schulman2017proximal}
    & \textbf{99.9} & $-$ & 192.1\ci{6.5} & 190.5\ci{8.0} & 13.8\ci{3.4}
    & 1.1 & 10.9\ci{1.5} & 166.1\ci{34.9} & 165.0\ci{21.7} & 51.8\ci{8.8}
    & 0.0 & $-$ & 193.1\ci{38.6} & 188.0\ci{29.3} & 52.3\ci{9.1}
    \\
    AMP~\cite{peng2021amp}
      & 74.2 & 5.5\ci{2.3} & 119.8\ci{20.9} & 138.9\ci{30.6} & 36.5\ci{7.5}
      & 94.5 & 4.3\ci{2.2} & 197.9\ci{11.9} & 222.5\ci{34.8} & 62.1\ci{10.8}
      & 95.1 & 3.3\ci{1.8} & 271.7\ci{64.2} & 265.0\ci{48.0} & 58.5\ci{38.2}
      \\
    AdaMimic~\cite{huang2025towards}
      & 0.0 & $-$ & 54.3\ci{6.1} & 63.2\ci{15.8} & 63.6\ci{73.4}
      & 0.0 & $-$ & 67.4\ci{8.2} & 63.5\ci{18.6} & 55.8\ci{6.9}
      & 4.7 & 12.1\ci{3.0} & 80.1\ci{24.7} & 86.9\ci{47.2} & 88.0\ci{90.2}
      \\
    DeepMimic~\cite{peng2018deepmimic}
      & 99.7 & \textbf{2.1}\ci{1.2} & 73.8\ci{3.9} & 79.3\ci{9.0} & \textbf{27.7}\ci{3.4}
      & \textbf{100.0} & \textbf{1.9}\ci{0.9} & 107.7\ci{2.8} & 268.7\ci{8.1} & 51.2\ci{5.1}
      & \textbf{100.0} & 2.4\ci{1.4} & 70.2\ci{3.7} & 70.1\ci{5.6} & 35.5\ci{4.3}
      \\
    DeepMimic-Distill
      & 99.3 & 2.2\ci{1.3} & 60.6\ci{4.6} & 64.3\ci{6.8} & 27.5\ci{4.7}
      & \textbf{100.0} & 2.1\ci{1.3} & 74.0\ci{3.4} & 64.0\ci{4.7} & 44.7\ci{4.7}
      & \textbf{100.0} & \textbf{2.1}\ci{1.2} & 60.3\ci{3.9} & 58.9\ci{4.9} & \textbf{34.8}\ci{4.1}
      \\
    \ours-w/o-Planner
      & 99.4 & 2.3\ci{1.5} & 62.0\ci{3.5} & 60.6\ci{5.8} & 29.0\ci{3.5}
      & 99.9 & 2.0\ci{1.2} & 66.4\ci{3.7} & 62.2\ci{6.0} & 46.7\ci{5.6}
      & \textbf{100.0} & 2.6\ci{1.5} & 70.7\ci{3.5} & 78.6\ci{8.8} & 37.5\ci{4.3}
      \\
    \ourrow\textbf{\ours (ours)}
      & 99.7 & 2.2\ci{1.2} & \textbf{58.1}\ci{5.8} & \textbf{59.9}\ci{7.3} & \textbf{25.2}\ci{6.8}
      & 99.9 & 2.3\ci{1.4} & \textbf{66.2}\ci{3.7} & \textbf{60.6}\ci{5.7} & \textbf{42.0}\ci{4.6}
      & 99.9 & 2.4\ci{1.4} & \textbf{53.4}\ci{4.4} & \textbf{54.3}\ci{6.8} & 35.8\ci{4.8}
      \\
    \bottomrule
  \end{tabular}
  \end{threeparttable}
  }
  \label{tab:main_simulation_results}
  \vspace{-0.2in}
\end{table*}

\subsection{Training Details}\label{subsec:training-details}

All policies are trained in Mjlab~\cite{zakka2026mjlab} using PPO~\cite{schulman2017proximal} with 4096 parallel simulation environments on 4 NVIDIA RTX 4090 GPUs. Both the policy and value networks are parameterized as 3-layer MLPs. We present more details, such as reward and MVAE states in the Appendix.


\subsection{Real-world Deployment}\label{subsec:deployment}
\paragraph{Ball localization.}
We consider two ball perception setups for real-world deployment. In the motion-capture setup, we directly obtain the 3D ball position from markers attached to the ball. In the camera-based setup, we use a YOLO26~\cite{redmon2016you} detector followed by stereo triangulation to estimate the 3D ball position, rather than relying on hand-crafted HSV-based color segmentation and region-of-interest ~\cite{zaidi2023athletic}. We find our solution is more robust and accurate. For rallying, we use a 6-camera stereo vision system following ESTHER~\cite{zaidi2023athletic}, while a single stereo camera is sufficient for serving due to its more structured setup. Although we adopt a similar multi-view setup to ESTHER, we later analyze its limitations in our system in the experiments.

\paragraphbegin{Future trajectory prediction.} Given the estimated 3D ball position and velocity, we predict future ball trajectories using a simple physics-based model that ignores spin and considers only gravity and aerodynamic drag. We propagate the trajectory forward under a constant drag model to obtain short-horizon predictions. 

\paragraphbegin{Robot localization.}
For robot localization, we use MoCap in controlled laboratory experiments and HTC VIVE Ultimate Trackers~\cite{HTC2024VIVEUltimateTracker} for in-the-wild deployment. The motion capture system provides accurate full-body pose estimation, while the VIVE trackers provide the robot's global pose without requiring a MoCap environment, enabling deployment outside the laboratory.

\paragraphbegin{High-level policy switching.}
We find that a single rally policy can suffer from motion drift when no ball is present: the autoregressive MVAE may gradually deviate from its stationary state, causing undesired oscillations. To address this issue, we employ two high-level policies: a standing policy and a rally policy. The standing policy maintains a stable posture when the robot is not actively responding to the ball, while the rally policy generates task-specific motions for ball interception and hitting. We switch between the two policies according to the current ball state.

%% file: sec/experiments.tex
\section{Experiments}
\subsection{Experimental Setup}

\begin{figure}[h]
    \centering
    \vspace{-0.12in}
    \includegraphics[width=1\linewidth]{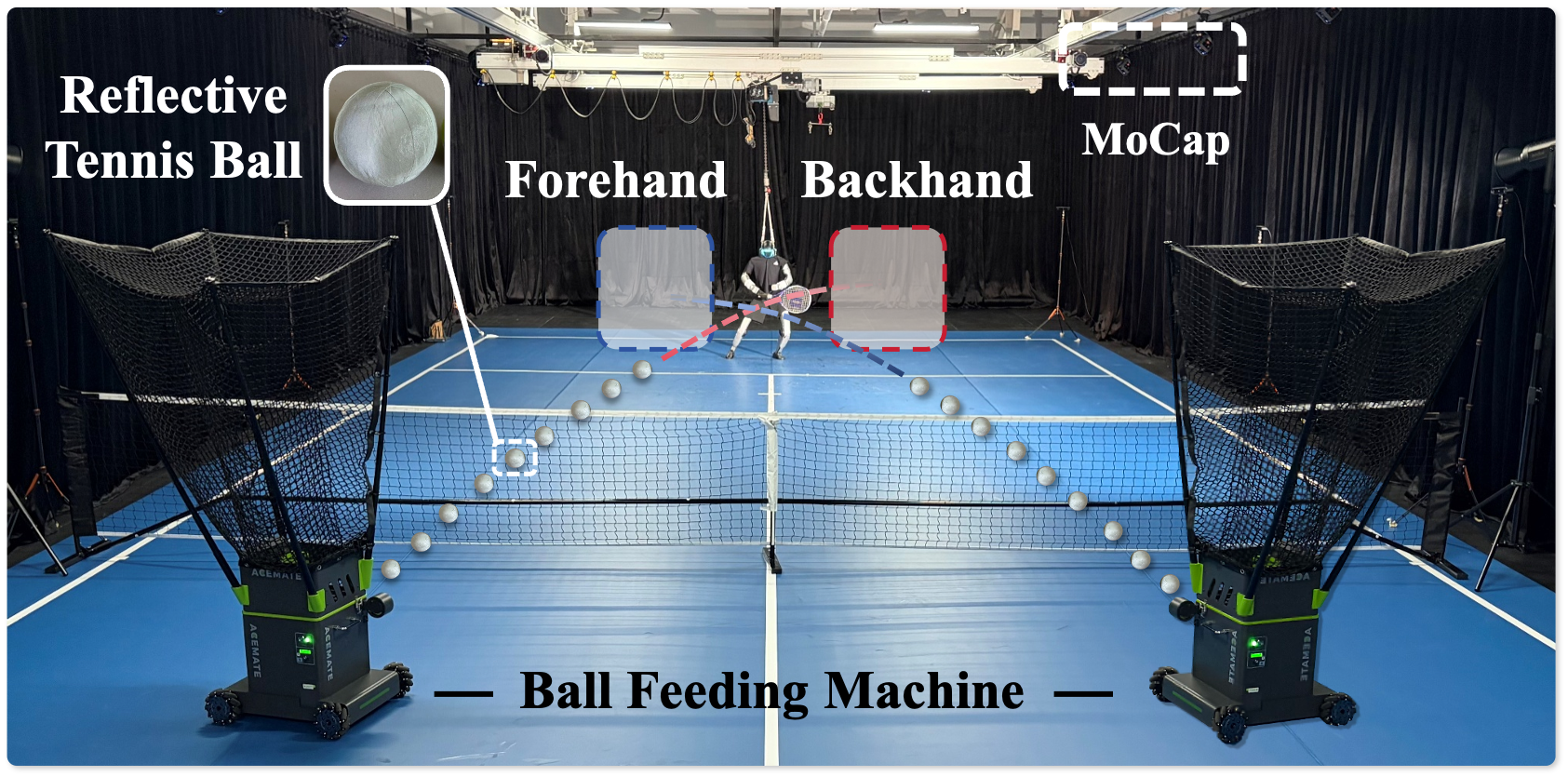}
    \vspace{-0.23in}
    \caption{Experimental setup for rally and serve task with motion capture.}
    \vspace{-0.12in}
    \label{fig:analyze_adapter}
\end{figure}

\paragraph{Hardware devices.} The main experiments are conducted in an approximately $20\mathrm{m} \times 8\mathrm{m}$\, motion-capture arena equipped with 35 Noitom MCC-400 cameras~\cite{noitom} for tracking the robot root pose and ball position. We use the Unitree G1 robot for testing. The tennis ball is customized to be reflective by attaching reflective materials to standard tennis balls. Due to the limited capture space, we adopt the ‘Tennis 10s’ system, whose singles court size is approximately $18 \times 6.5$\,m. This setting is also well-matched to the physical capabilities of G1, with a height of 1.35 m. We use a 23-inch junior tennis racket with a wrist mount fixed at a 25-degree offset to approximate a natural racket-gripping posture. \cref{fig:analyze_adapter} visualizes the experiment setup. 

\paragraphbegin{Evaluation.} For fair rally evaluation in real robot experiments, we use an Acemate ball feeding machine~\cite{acemate}, and symmetrically place it to launch balls into predefined target regions with reproducible trajectories, repeating 25 times for each region. These launcher configurations are unseen during training, enabling reliable evaluation of forehand and backhand generalization. Each trial launches one ball every 4.5s. For serving evaluation, we use the Synria Gloria-D parallel gripper~\cite{synria} to grasp and toss the tennis ball with 15 trials per model.  We also evaluate the Unitree Dex-3 dexterous hand on simpler amateur-level motions; its weight (much heavier than the parallel gripper) prevents G1 from balancing under professional serving motions.

\paragraphbegin{Metrics.} We employ three shared metrics for both tasks: hit success rate, bounce-position error $E_{\mathrm{bo}}$, and joint acceleration error $E_{\mathrm{acc}}$, which assess motion smoothness. The target position is fixed at the center of the opponent's court for rally shots, and randomly sampled inside the diagonal service box as a policy command for serves. For rally, we adopt the Fr\'echet Inception Distance $E_{\mathrm{FID}}$~\cite{heusel2017gans} in joint space to measure style similarity to reference motions, alongside the power metric $E_{\mathrm{pow}}$ for motion intensity. For serve, we present global and local Mean Per-Keyframe Position Error (MPKPE), namely $E^{\mathrm{global}}_{\mathrm{mpkpe}}$ and $E^{\mathrm{local}}_{\mathrm{mpkpe}}$.

\paragraphbegin{Baselines.} We employ several shared baselines across both tasks. RL-Scratch (with PPO~\cite{schulman2017proximal}) learns the policy without motion data, while AMP~\cite{peng2021amp} incorporates adversarial style rewards from reference motions. For rally, we also compare against large-scale motion-prior approaches, including the continuous-latent-prior PULSE~\cite{luo2024universal} and the discrete-prior NCP~\cite{zhu2023neural,han2024lifelike}. We additionally consider the recent LATENT~\cite{zhang2026learning} for humanoid tennis, which is built upon PULSE. Since its full implementation is not publicly available, we use PULSE as a proxy baseline to provide insight into it. Vid2Player3D~\cite{zhang2023learning} is a closely related simulation method that adopts a planner-tracker architecture without adaptation mechanisms. For serving, we include DeepMimic~\cite{peng2018deepmimic} and AdaMimic~\cite{huang2025towards}, where the latter extends single-reference imitation with speed adaptation. 

\paragraphbegin{Remark.} All subsequent experiments are evaluated using data from broadcast videos, while MoCap data is used only for real-world robot demonstrations.

\begin{table*}[tbp]
    \centering
    \setlength{\tabcolsep}{3pt}
    \renewcommand\arraystretch{1.0}
    \caption{\textbf{Real-robot evaluation of rally skills.} The results demonstrate that both the Adaptive Tracker and Adaptive Planner contribute to robust real-world rally performance over all three players. The performance gap between simulation and real-world deployment is further analyzed in~\cref{sec:analysis_sim2real}.}
    \vspace{-0.1in}
    \label{tab:real-rally-results}

    \resizebox{1\textwidth}{!}{
    \begin{tabular}{l|lcc|ccc|ccc|ccc|ccc}
    \toprule

    \multirow{2}{*}{\textbf{Player}} &
    \multirow{2}{*}{\textbf{Method}} &
    \multirow{2}{*}{\shortstack{Adaptive\\Tracker}} &
    \multirow{2}{*}{\shortstack{Adaptive\\Planner}} &

    \multicolumn{3}{c|}{Hit Rate (\%) $\uparrow$} &
    \multicolumn{3}{c|}{Net Clearance Rate (\%) $\uparrow$} &
    \multicolumn{3}{c|}{$E_{\mathrm{FID}}$ $\downarrow$} &
    \multicolumn{3}{c}{$E_{\mathrm{acc}}$ $\downarrow$} \\

    \cmidrule(lr){5-7}
    \cmidrule(lr){8-10}
    \cmidrule(lr){11-13}
    \cmidrule(lr){14-16}

    & & & &
    Sim. & Real-FH & Real-BH &
    Sim. & Real-FH & Real-BH &
    Sim. & Real-FH & Real-BH & 
    Sim. & Real-FH & Real-BH \\
    \midrule

    \multirow{4}{*}{\textbf{\nadalfull}}
    & Vid2Player3D & $\thickcdot$ & $\thickcdot$ 
    & 82.6 & 28.0 & 28.0 
    & 68.5 & 28.5 & 14.8
    & 7.9 & 15.4 & \textbf{10.4}
    & 45.2 & 18.8 & 14.7 \\
    & \textbf{\ours (ours)} & \Checkmark & \Checkmark 
    & \textbf{91.5} & \textbf{44.0} & \textbf{56.0}  
    & \textbf{72.8} & 27.2 & 35.7
    & \textbf{5.3} & 19.4 & 12.9
    & 43.0 & 19.9 & 16.3 \\
    & w/o-AdaPlanner & \Checkmark & $\thickcdot$ 
    & 71.7 & \textbf{44.0} & 28.0 
    & 42.1 & 27.2 & 28.5
    & 5.5 & 22.2 & 14.3
    & \textbf{40.8} & \textbf{16.2} & \textbf{14.5} \\
    & w/o-AdaTracker & $\thickcdot$ & \Checkmark 
    & 80.6 & 40.0 & 36.0
    & 62.9 & \textbf{50.0} & \textbf{44.4} 
    & 8.0 & \textbf{18.3} & 12.9
    & 45.7 & 19.2 & 19.1 \\
    \midrule

    \multirow{4}{*}{\textbf{\federerfull}}
    & Vid2Player3D & $\thickcdot$ & $\thickcdot$ 
    & 81.5 & 24.0 & 8.0 
    & \textbf{88.3} & 16.6 & 0.0 
    & \textbf{5.1} & \textbf{9.1} & \textbf{9.2} 
    & 41.1 & 16.5 & \textbf{15.3}\\
    & \textbf{\ours (ours)} & \Checkmark & \Checkmark 
    & \textbf{96.3} & \textbf{64.0} & \textbf{48.0} 
    & 86.9 & \textbf{56.3} & \textbf{50.0} 
    & 5.4 & 15.1 & 13.9 
    & \textbf{16.3} & \textbf{14.9} & 21.6 \\
    & w/o-AdaPlanner & \Checkmark & $\thickcdot$  
    & 80.9 & 52.0 & 16.0  
    & 67.5 & 15.3 & 0.0 
    & 10.7 & 18.8 & 17.9 
    & 41.3 & 17.0 & 15.9\\
    & w/o-AdaTracker & $\thickcdot$ & \Checkmark 
    & 87.5 & 48.0 & 20.0 
    & 86.5 & 33.3 & 20.0 
    & 8.6 & 14.2 & 12.1 
    & 47.7 & 19.9 & 23.2 \\
    \midrule

    \multirow{4}{*}{\textbf{\djokovicfull}}
    & Vid2Player3D & $\thickcdot$ & $\thickcdot$ 
    & 83.2 & 8.0 & 20.0
    & 70.0 & 0.0 & 20.0
    & 6.3 & 17.7 & 13.7
    & 40.6 & 16.1 & 19.4 \\
    & \textbf{\ours (ours)} & \Checkmark & \Checkmark 
    & \textbf{92.3} & \textbf{48.0} & \textbf{48.0}  
    & 75.6 & \textbf{33.3} & \textbf{33.3} 
    & 6.3 & \textbf{12.0} & 11.8
    & \textbf{41.4} & 19.6 & 18.2 \\
    & w/o-AdaPlanner & \Checkmark & $\thickcdot$ 
    & 86.4 & 32.0 & 20.0 
    & 55.5 & 25.0 & 20.0 
    & \textbf{6.2} & 19.5 & 14.6 
    & 41.5 & 15.4 & \textbf{16.7} \\
    & w/o-AdaTracker & $\thickcdot$ & \Checkmark 
    & 83.9 & 24.0 & 24.0 
    & \textbf{78.5} & 16.7 & \textbf{33.3} 
    & 12.0 & 12.8 & \textbf{11.1} 
    & 42.5 & \textbf{11.0} & 20.3 \\
    \bottomrule
    \end{tabular}
    }
    \vspace{-0.15in}
\end{table*}

\subsection{Simulation Results and Analyses}\label{sec:sim-results}
We present the main simulation results in~\cref{tab:main_simulation_results} and analyze different methods under rally and serve settings, aiming to extract key insights on the algorithmic side.

\begin{wrapfigure}{r}{0.42\linewidth}
    \centering
    \vspace{-0.17in}

    \includegraphics[width=\linewidth]{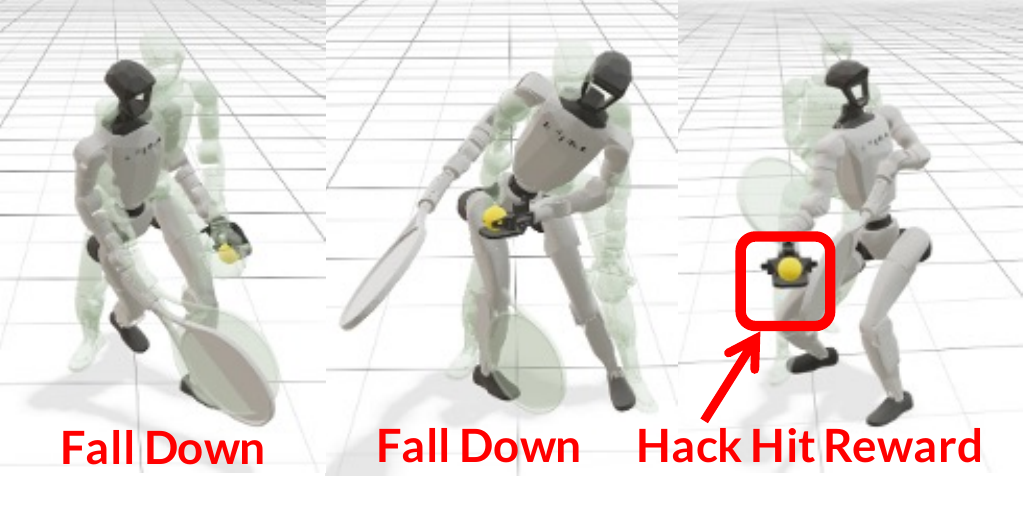}
    \vspace{-0.3in}
    \caption{Analysis of failure of RL-Scratch serving.}
    \label{fig:rlscratch}
    \vspace{-0.2in}
    
\end{wrapfigure}
\paragraphbegin{RL-Scratch verifies the importance of motion data.}
In rally, it achieves high hit success, indicating strong reactive control capability in continuous ball interactions. However, its generated
motions exhibit large deviations from human-like stroke dynamics, leading to high FID. This gap
becomes more pronounced in serve, where precise ball placement and stable whole-body coordination are required. In this setting, RL-Scratch succeeds for only one player style. This reflects a
strong sensitivity to player-specific serving strategies, where different players exhibit distinct ball-tossing trajectories and pre-serve motion patterns. Specifically,
the narrow striking window provides little RL exploration guidance, while unstable postures can prevent basic balance. The Nadal case is an exception: its stable
initial posture enables balance and reward hacking by keeping the
racket attached to the ball (\cref{fig:rlscratch}).

\begin{table}[t]
    \centering
    \setlength{\tabcolsep}{2.5pt} %
    \renewcommand\arraystretch{1.05} %
    \caption{\textbf{Real-robot evaluation of serve skills.} The results demonstrate that both the Adaptive Tracker and Adaptive Planner contribute to robust real-world rally performance over all three players.
Performance drops between simulation and real-world deployment are discussed in~\cref{sec:analysis_sim2real}.}
    \label{tab:real-serve-results}
    \vspace{-0.1in}
    \resizebox{1\linewidth}{!}{
    \begin{tabular}{l|l|cccccc}
    \toprule
    \textbf{Player} & \textbf{Method} & Succ. $\uparrow$ &  $E_{\mathrm{Bo}}^{\mathrm{3m}} \downarrow$ & $E_{\mathrm{Bo}}^{\mathrm{6m}} \downarrow$ & $E_{\mathrm{Bo}}^{\mathrm{9m}} \downarrow$ & $E_{\mathrm{dof}}^{\mathrm{track}} \downarrow$ & $E_{\mathrm{acc}} \downarrow$ \\
    \midrule
    
    \multirow{4}{*}{\textbf{\nadalfull}}  
    & DeepMimic & 80.0 & 0.7\ci{0.3} & 3.5\ci{0.1} & 6.4\ci{0.6} & 0.18\ci{0.01} & 9.9\ci{2.9} \\
    &  \textbf{\ours (ours)} & 66.7 & 1.5\ci{0.5} & 1.5\ci{0.4} & 5.3\ci{1.7} & 0.17\ci{0.01} & 12.3\ci{0.5} \\
    & w/o-Planner & 53.3 & 1.8\ci{1.1} & 2.8\ci{0.9} & 4.6\ci{1.7} & 0.18\ci{0.01} & 13.5\ci{0.2} \\
    & w/o-TossRew & 60.0 & 1.3\ci{0.4} & 2.5\ci{1.4} & -- & 0.18\ci{0.01} & 11.7\ci{0.7} \\
    \midrule
    
    \multirow{4}{*}{\textbf{\federerfull}} 
    & DeepMimic & 53.3 & 1.6\ci{1.0} & 4.2\ci{1.5} & 7.8\ci{0.0} & 0.36\ci{0.03} & 28.4\ci{0.9} \\
    &  \textbf{\ours (ours)}  & 73.3 & 1.4\ci{1.7} & 2.8\ci{1.2} & 6.4\ci{1.4} & 0.17\ci{0.01} & 22.0\ci{1.0} \\
    & w/o-Planner & 53.3 & 1.1\ci{0.2} & 4.0\ci{0.8} & 5.1\ci{1.0} & 0.21\ci{0.02} & 29.8\ci{5.3} \\
    & w/o-TossRew & 73.3 & 1.2\ci{0.7} & 3.4\ci{0.6} & 7.0\ci{0.6} & 0.18\ci{0.02} & 29.2\ci{5.9} \\
    \midrule
    
    \multirow{4}{*}{\textbf{\djokovicfull}} 
    & DeepMimic & 60.0 & 2.5\ci{0.00} & 4.2\ci{0.00} & -- & 0.23\ci{0.01} & 19.1\ci{2.0} \\
    &  \textbf{\ours (ours)} & 86.7 & 1.8\ci{0.8} & 2.0\ci{1.3} & 5.6\ci{1.7} & 0.14\ci{0.02} & 22.8\ci{8.0} \\
    & w/o-Planner & 53.3 & 2.5\ci{0.4} & 2.3\ci{0.8} & 7.6\ci{0.1} & 0.17\ci{0.02} & 22.7\ci{6.6} \\
    & w/o-TossRew & 80.0 & 1.6\ci{0.2} & 3.8\ci{0.5} & 6.5\ci{1.1} & 0.18\ci{0.01} & 19.0\ci{6.7} \\
    \bottomrule
    
    \end{tabular}
    }
    \vspace{-0.15in} %
\end{table}

\paragraphbegin{AMP struggles at capturing phase and whole-body coordination.}
In rally, AMP produces more
human-like stroke motions, but footwork remains highly sensitive to task objectives and is often
distorted by ball recovery demands, leading to significantly reduced whole-body coordination across
player styles. In serving, this limitation is amplified: without explicit phase structure, the model
fails to capture the full force-generation process from preparation to strike~\cite{huang2025towards}. As a result, AMP
can underperform RL-Scratch despite style rewards, suggesting that adversarial imitation alone is
insufficient to capture perceptive, phase-consistent, and whole-body coordination in ball sports.

\paragraphbegin{Decoupled planning and tracking improves rally styles.}
Motion-prior methods such as PULSE and NCP, despite their hierarchical structure, still exhibit strong coupling between planning and tracking. 
In particular, their VAE-based encoders condition on both the motion reference and the current robot proprioception, causing the latent representation to be heavily entangled with robot state. This limits the expressiveness of the motion prior,
leading to higher $E_{\mathrm{FID}}$, while the coupling enables more responsive control.
This coupling is also reflected in physically noticeable artifacts, including distorted swing motions and inconsistent footwork patterns during ball recovery, suggesting a loss of stylistic motion under interaction-driven execution. 
In contrast, Vid2Player3D explicitly decouples planning and tracking, which improves stylistic consistency and results in significantly better $E_{\mathrm{FID}}$, while introducing a degradation in task performance due to reduced responsiveness to interaction dynamics. To address this trade-off, our proposed speed-adaptive mechanism better balances between task responsiveness and styles in the rally task.

\paragraphbegin{Introducing phase-aware speed adaptation improves serving style.}
Directly applying DeepMimic with task conditioning achieves reasonable serve success, but often sacrifices stylistic details such as wrist rotation and follow-through due to coupling between task reward and motion style~\cite{Hitter2025,ren2026smash}. While AdaMimic enables joint optimization of motion speed, tracking, and task execution, full-trajectory optimization leads to unstable learning and suboptimal convergence in the serving task. \ours addresses these issues via adaptation and a phase mask.

\subsection{Real-world Results and Analyses}\label{sec:real-results}

In real-world evaluation, ceiling collisions may occur for high-arcing rally shots, making estimation of landing positions unreliable. Instead, we report the net clearance rate in~\cref{tab:real-rally-results}. Serves are more controlled in height, enabling reliable evaluation of landing accuracy in~\cref{tab:real-serve-results}.

\paragraphbegin{Significant sim-to-real gap of planning-tracking architecture.}
Directly applying Vid2Player3D leads to a significant sim-to-real degradation in rally performance (\cref{tab:real-rally-results}). 
While it achieves reasonable results in simulation, its real-robot performance drops substantially across both forehand and backhand settings, especially in net clearance rate and hitting consistency. 
This highlights the limitation of the decoupled planner and tracker in real-world dynamics.

\paragraphbegin{Effect of adaptive tracking and planning in rally task.} Both adaptive tracking and planning improve real-robot rally performance by addressing different sim-to-real gaps. The adaptive tracker reduces low-level execution errors, yielding consistent gains over the non-adaptive variant across forehand and backhand settings. In contrast, the adaptive planner improves high-level temporal coordination by adjusting execution speed, leading to higher hit rates compared to Vid2Player3D. However, each component alone is insufficient: tracking alone cannot resolve timing errors, while planning alone remains sensitive to control noise. Combining both yields the best performance, demonstrating their complementarity in mitigating both temporal and execution-level errors.

\begin{figure}[t]
    \centering
    \includegraphics[width=1\linewidth]{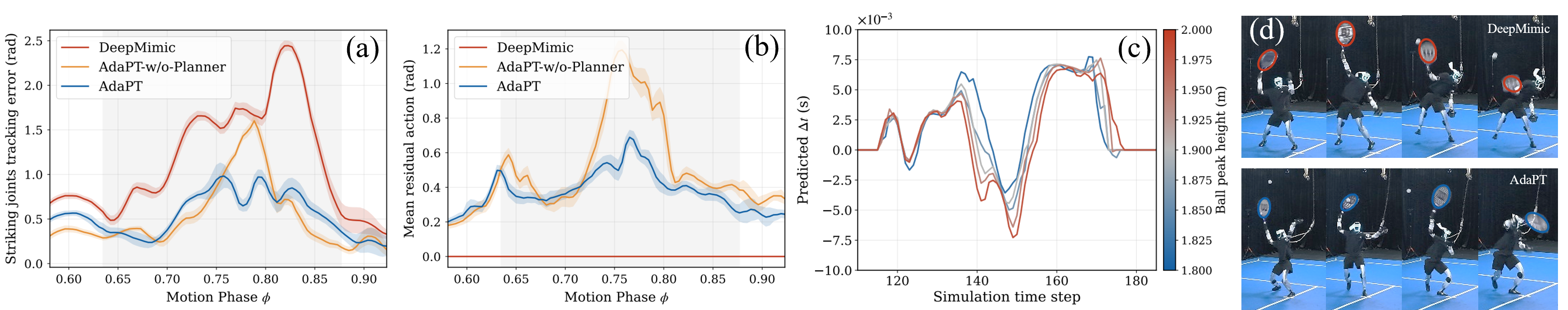}
    \caption{\textbf{Planner analysis.} The planners successfully learn to adjust motion speed to (a, b) minimize tracking error and (c) adapt to varied ball-tossing height, (d) exhibiting better style than DeepMimic.}
    \vspace{-0.2in}
    \label{fig:analyze_dt}
\end{figure}

\paragraphbegin{Effect of speed adaptation on stylized serving.}
Compared with DeepMimic, our two-stage serve training better preserves player-specific motion styles and achieves lower tracking errors (\cref{tab:real-serve-results}). However, directly tracking stylized motions is insufficient for robust real-world serving, as fixed execution timing can degrade
task accuracy under disturbances. 
Our adaptive mechanism addresses this by adjusting execution speed during deployment, enabling faster racket acceleration and stronger long-distance serves (low $E_{\mathrm{bo}}^{\mathrm{9m}}$ error), while preserving realistic wrist pronation and player-specific serving styles (\cref{fig:analyze_adapter}).

\subsection{Analyses of Key Components}\label{sec:analysis}
We analyze several key components in our system that are critical for learning professional tennis skills, providing insights for humanoid ball sports in the future.

\begin{wrapfigure}{r}{0.44\linewidth}
    \centering
    \vspace{-0.15in}
    \includegraphics[width=1\linewidth]{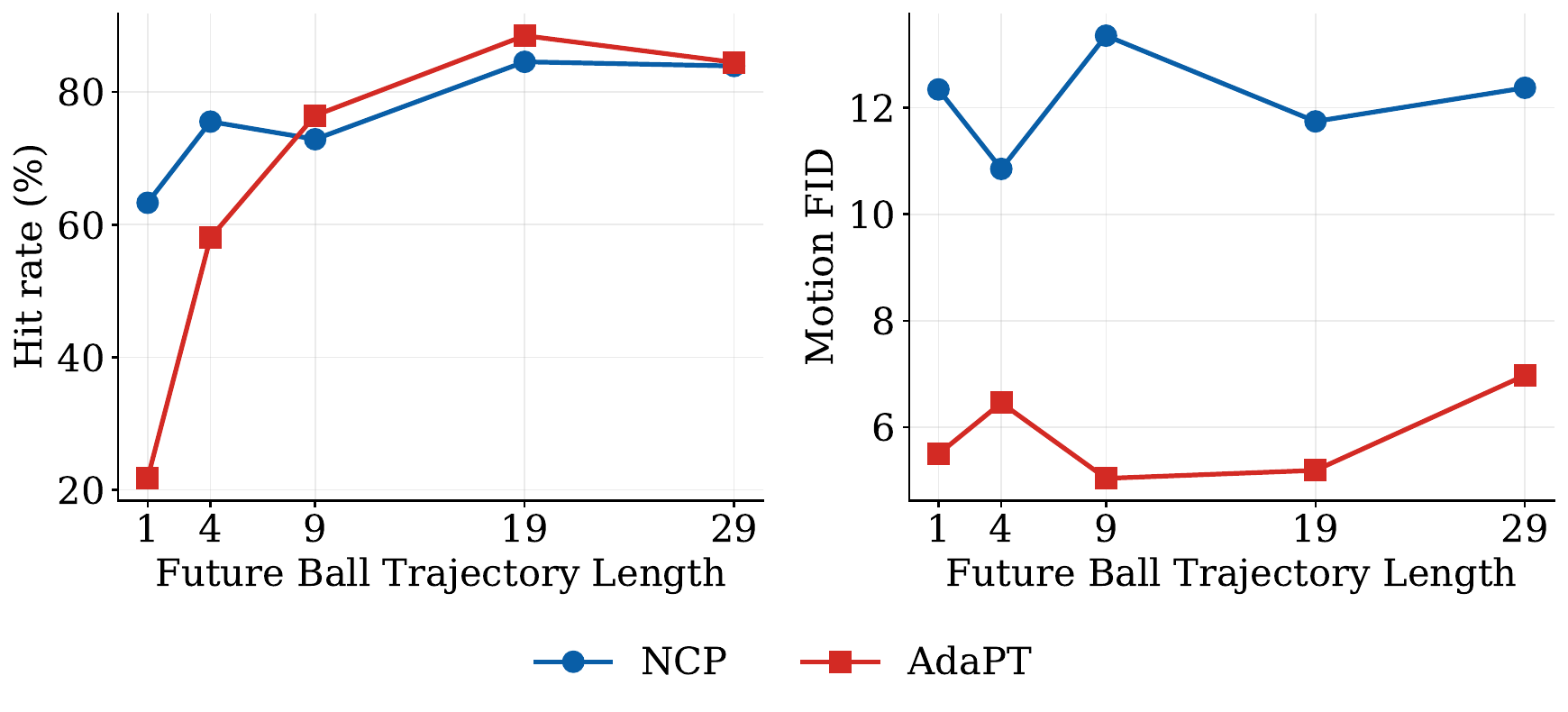}
    \vspace{-0.25in}
    \caption{Sensitivity of planner-tracker methods to the horizon of the future ball trajectory.}
    \label{fig:tradeoff-analysis}
    \vspace{-0.2in}
\end{wrapfigure}
\paragraphbegin{Trade-off between rally responsiveness and motion style fidelity.}
We find that incorporating future ball trajectory observations consistently improves task performance in both simulation and the real world. We attribute this to the decoupled planner-tracker architecture, where the planner benefits from longer-horizon information for improved foresight in trajectory planning. This observation motivates the hypothesis that tightly coupled policies, such as NCP and PULSE, may rely less on long-horizon observations due to their stronger real-time responsiveness. To validate this, we evaluate different future observation horizons for both types of methods in \cref{fig:tradeoff-analysis}. It confirms a key trade-off:
\vspace{-0.02in}

\takeaway{Decoupled planning-tracking architectures rely on longer and more accurate lookahead ball predictions to preserve temporally coherent motion styles. In contrast, coupled architectures, such as NCP- or PULSE-style policies, react directly to current observations and offer better real-time responsiveness, often at the cost of motion naturalness.}
\vspace{-0.02in}
\noindent This finding provides a possible explanation for why observing only the current ball position and velocity is already sufficient for learning rally in LATENT~\cite{zhang2026learning}, which is built upon a PULSE-style tightly coupled architecture.

\paragraphbegin{Adaptation works for serve.}
We analyze tracking errors and residual actions across motion phases in \cref{fig:analyze_dt}. 
AdaPT achieves the lowest tracking error during the hitting phase with smaller residual actions.
Meanwhile, the planner dynamically adjusts the motion speed according to the ball-toss height, enabling adaptation to varying tosses.

\begin{wrapfigure}{r}{0.45\linewidth}
    \centering
    \vspace{-0.2in}
    \includegraphics[width=1\linewidth]{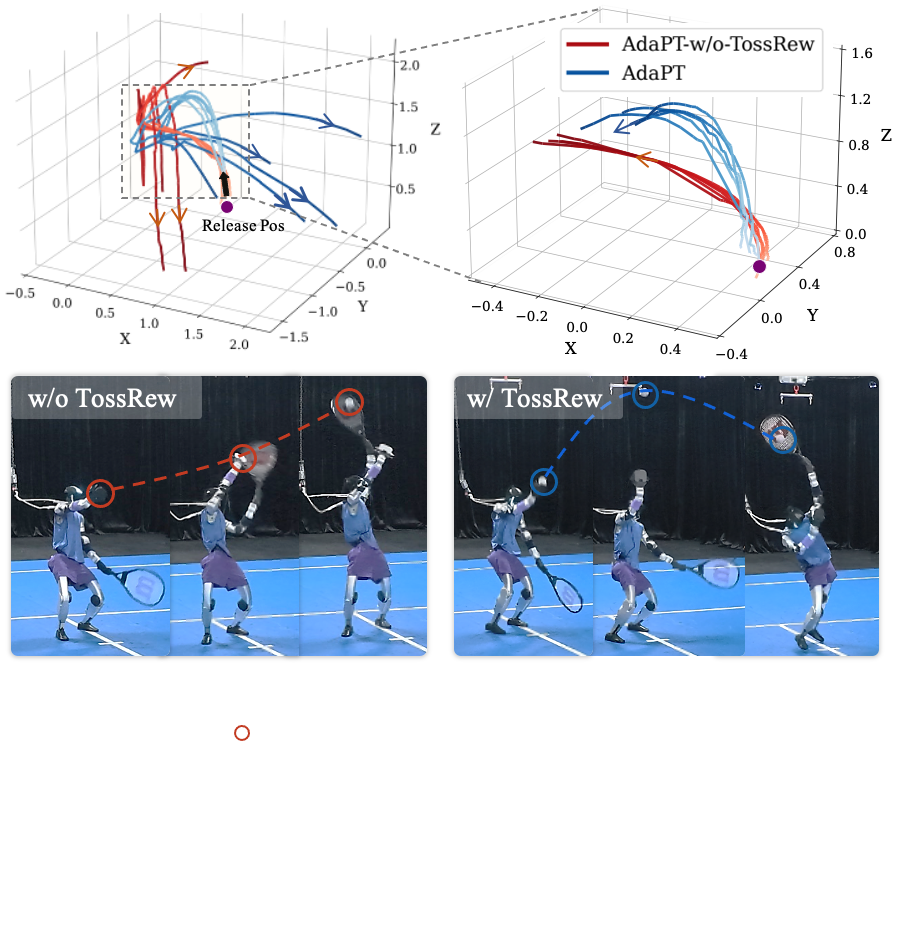}
    \vspace{-0.72in}
    \caption{\textbf{Tossing guidance} as a reward is crucial to imitate tossing style and control landing.}
    \label{fig:toss_reward}
    \vspace{-0.2in}
\end{wrapfigure}
\paragraphbegin{Tossing guidance is important for serving.}
Interestingly, we observe that, without tossing guidance, the model can still achieve reasonable serve success rates, but exhibits significantly worse control over ball landing positions (see \cref{tab:real-serve-results}). To better understand this phenomenon, we analyze real-robot serving trajectories and find that, without tossing guidance, the policy tends to produce overly low tosses in order to simplify ball interception and improve stability. However, this leads to insufficient reaction time and limited controllability over the final landing location.
In contrast, introducing tossing guidance encourages a higher and more consistent toss trajectory. This not only provides the robot with more preparation time but also enables stronger racket acceleration and more effective “whipping” motions. As a result, the robot can better reproduce fine-grained technical details such as wrist pronation, while achieving more accurate control of the serve landing position.

\subsection{Analyses of Sim-to-real Gap}\label{sec:analysis_sim2real}

\begin{figure}[h]
    \centering
    \vspace{-0.1in}
    \includegraphics[width=\linewidth]{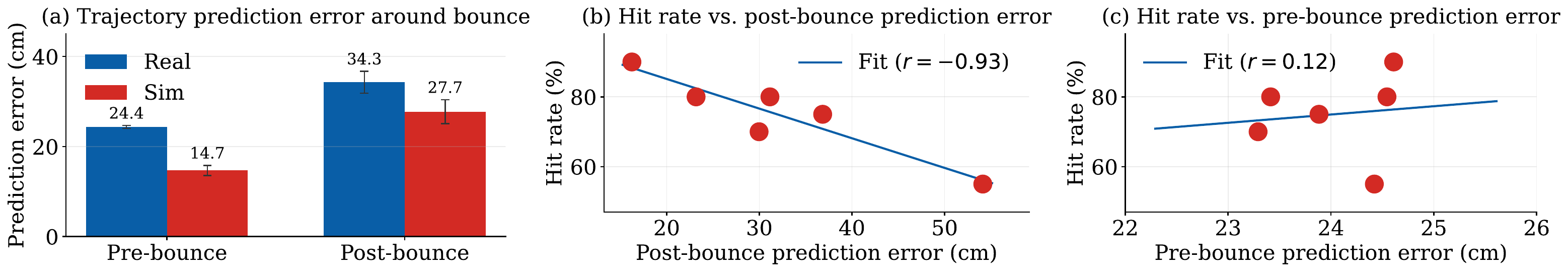}
    \vspace{-0.27in}
\caption{\textbf{Impact of trajectory prediction errors on rally success.}
(a) Future ball trajectory prediction errors. (b,c) Correlation between pre- and post-bounce prediction errors and rally success, respectively.}
\vspace{-0.2in}
\label{fig:ball-pred-error}
\end{figure}

\paragraph{Post-bounce trajectory prediction is critical for rally.}
Future ball trajectory prediction is affected by inevitable velocity estimation noise and irregular ball-ground interactions caused by the ball's reflective surface, particularly after the bounce (\cref{fig:ball-pred-error}a). 
More importantly, post-bounce prediction errors exhibit a stronger negative correlation with rally success than pre-bounce errors (\cref{fig:ball-pred-error}b,c), indicating that accurate post-bounce trajectory prediction is particularly important for reliable ball interception.

\begin{figure}[h]
    \centering
    \begin{minipage}[t]{0.52\linewidth}
        \vspace{-0.02in}
        \centering
        \resizebox{\linewidth}{!}{
        \begin{tabular}{lccc}
            \toprule
            Method & $E_{\mathrm{dof}}^{\mathrm{track}}$ $\downarrow$ &
            $E_{\mathrm{MVAE}}^{\mathrm{FID}}$ $\downarrow$ &
            Succ. $\uparrow$ \\
            \midrule
            Vid2Player3D & 0.29 & 20.0 & 16\% \\
            AdaPT & 0.23 & \textbf{4.8} & \textbf{56\%} \\
            w/o AdaPlanner & \textbf{0.21} & 15.0 & 34\% \\
            w/o AdaTracker & 0.27 & 10.3 & 34\% \\
            \bottomrule
        \end{tabular}
        }
        \vspace{-0.02in}
        \captionof{table}{AdaPlanner and AdaTracker mitigate planning and tracking drift in the real robot experiments.}
        \label{tab:planning-tracking}
    \end{minipage}
    \hfill
    \vspace{-0.1in}
    \begin{minipage}[t]{0.45\linewidth}
        \vspace{0pt}
        \centering
        \includegraphics[width=\linewidth]{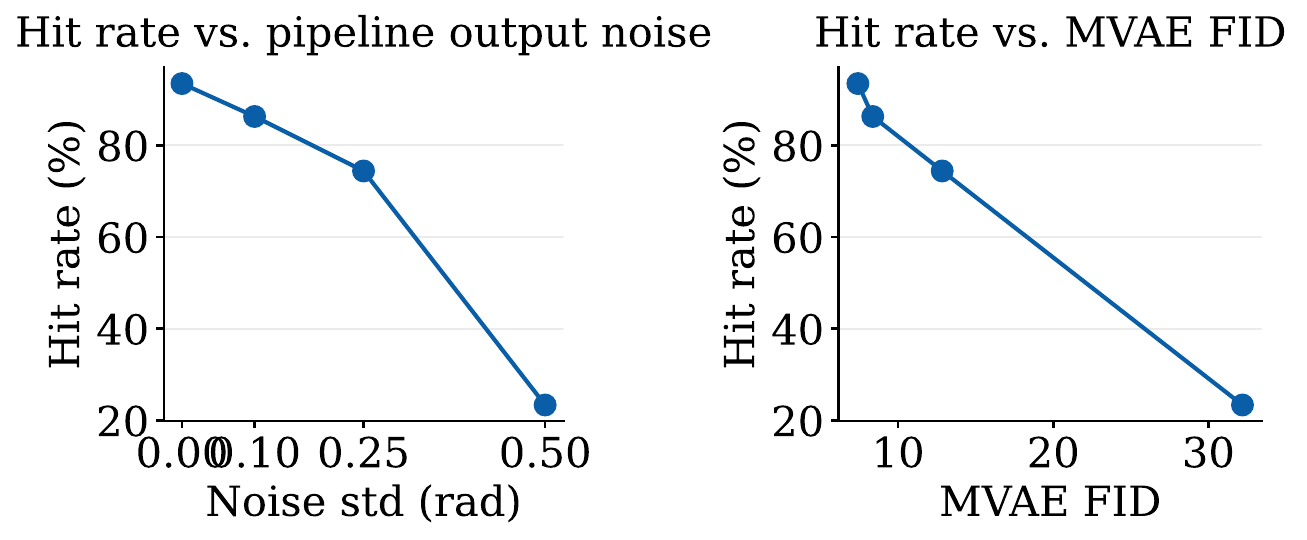}
        \vspace{-0.25in}
        \captionof{figure}{The negative correlation between motion fidelity and hit success in the simulation.}
        \label{fig:tracking-mae}
    \end{minipage}
    \vspace{-0.06in}
\end{figure}

\paragraphbegin{Planning and tracking drifts limit rally.}
Real-robot ablations show that AdaPlanner and AdaTracker respectively mitigate planning and tracking drift (\cref{tab:planning-tracking}). 
Simulation further reveals that both drifts are negatively correlated with hit success (\cref{fig:tracking-mae}), explaining the improved real-world performance of both single-component variants over Vid2Player3D and their further gains when combined.
\vspace{-0.2in}

\begin{figure}[!htbp]
    \centering
    \vspace{-0.14in}
    \begin{minipage}[c]{0.56\linewidth}
        \centering
        \setlength{\tabcolsep}{2.5pt}
        \resizebox{\linewidth}{!}{
        \begin{tabular}{l|ccccc}
            \toprule
            \textbf{Method} & Succ. $\uparrow$ & Fall. $\downarrow$ &
            $E_{\mathrm{Bo}}^{9m} \downarrow$ &
            $E_{\mathrm{dof}}^{\mathrm{track}} \downarrow$ &
            $E_{\mathrm{acc}} \downarrow$ \\
            \midrule
            w/o-DR & 20\% & 40.0 & 6.2
            & 0.18
            & 15.1
            \\
            w/o-BallDR & 40\% & 40.0 & 4.8
            & 0.17
            & 13.8
            \\
            w/o-PoseDR & 80\% & 20.0 & 5.3 
            & 0.17
            & 14.4
            \\
            \textbf{AdaPT}
            & \textbf{90\%} & \textbf{0.0} & \textbf{2.1}
            & \textbf{0.14}
            & \textbf{11.0}
            \\
            \bottomrule
        \end{tabular}
        }
        \captionof{table}{Effect of domain randomization on real-world serve robustness.}
        \label{table:dr_serve}
    \end{minipage}
    \hfill
    \begin{minipage}[c]{0.42\linewidth}
        \centering
        \vspace{0.05in}
        \includegraphics[width=\linewidth]{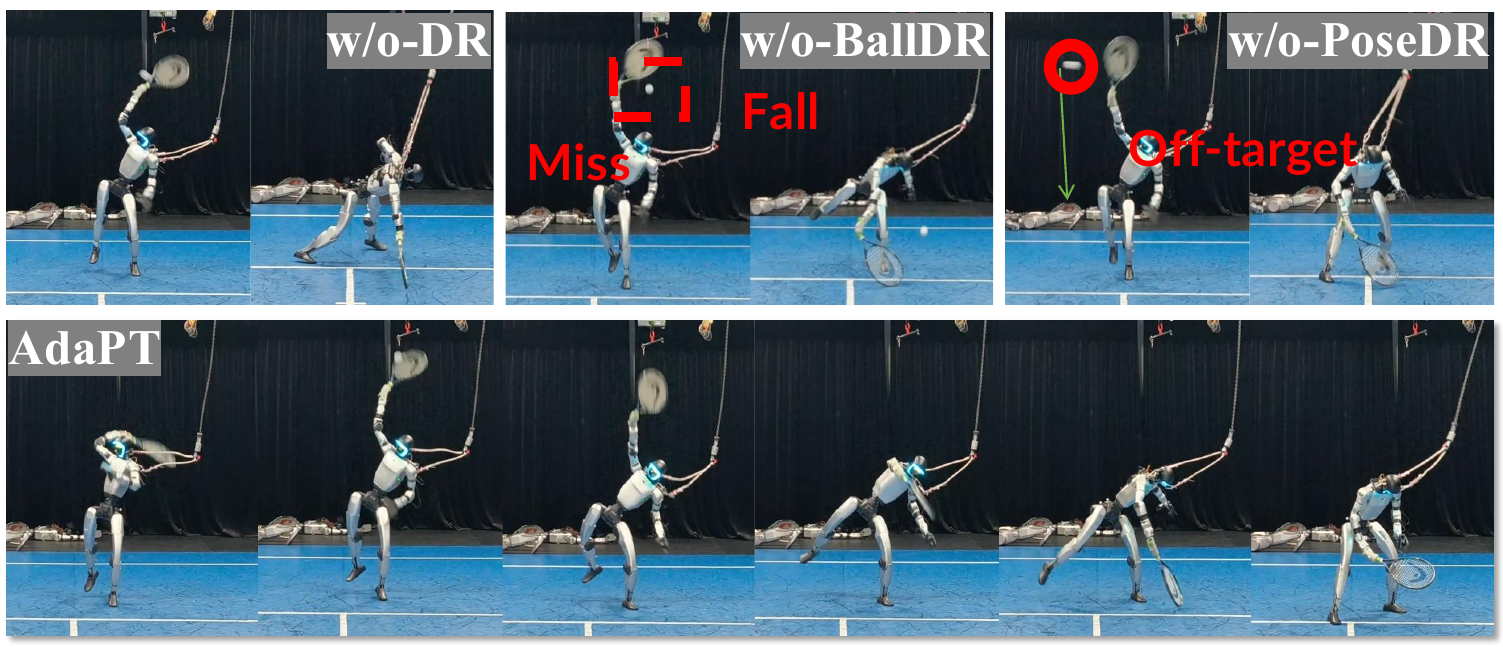}
            \captionof{figure}{Visualization of serve's domain randomization.}
        \label{fig:dr_serve}
    \end{minipage}
    \vspace{-0.2in}
\end{figure}

\paragraphbegin{Initial pose and ball perturbations improve serving robustness.}
We identify initial pose randomization and ball perturbation as important factors in reducing the sim-to-real gap for serving. 
Ball perturbation substantially improves success and prevents OOD-induced falls, while pose randomization reduces bounce error and improves robustness to variations in the initial orientation (\cref{table:dr_serve} and \cref{fig:dr_serve}).
\vspace{-0.1in}

\begin{figure}[h]
    \centering
    \vspace{-0.15in}
    \includegraphics[width=1\linewidth]{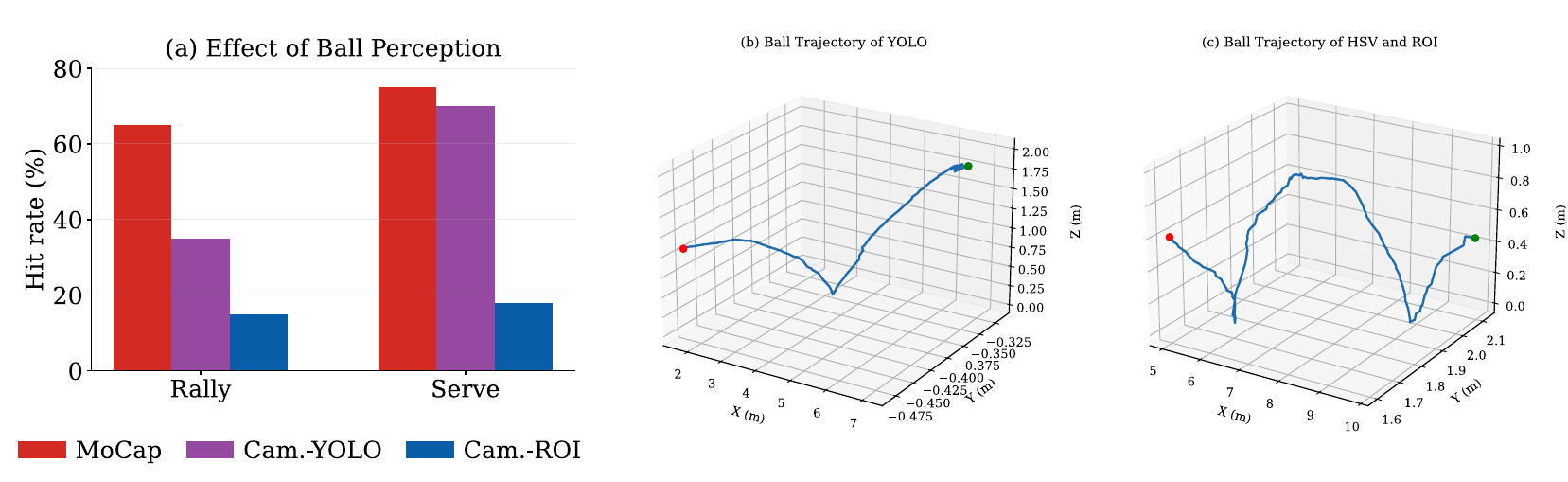}
    \vspace{-0.3in}
    \caption{Large trajectory estimation noise with camera influence on real performance. But YOLO-based perception is sufficient for serve. }
    \label{fig:camera_analysis}
    \vspace{-0.1in}
\end{figure}

\paragraphbegin{Large trajectory estimation noise with camera.} We further compare ball perception using the motion capture system and the multi-camera vision setup. Although the cameras are jointly calibrated, we observe larger localization noise and temporals jitter in vision-based detections (\cref{fig:camera_analysis}), which can lead to inaccurate trajectory prediction. We further evaluate two ball detection approaches and find that YOLO-based detection consistently outperforms HSV-based ball extraction with region-of-interest (ROI) filtering~\cite{zaidi2023athletic} for rallying, leading to a higher hit success rate. For service, YOLO achieves performance comparable to the motion capture system and substantially better than the HSV-based approach, likely because the cameras are positioned closer to the ball during the motion. Since our adaptation mechanism relies heavily on accurate future ball trajectory estimation, perception errors can substantially degrade real-world performance. These results suggest that camera-based perception remains a promising direction for scalable humanoid tennis systems, while highlighting the importance of accurate and temporally stable ball localization in real tennis court environments.

\vspace{-0.1in}
\section{Conclusion and Limitations}
We proposed \ours, a solution for stylized humanoid tennis playing. \ours learns from professional player motions and enables both rallying and serving. Through decoupled planning and tracking, \ours achieves strong task performance while preserving player-specific styles in real-world humanoid robots, G1 and Atom.

\paragraphbegin{Limitations.} The proposed system still relies primarily on motion capture to obtain the robot root and ball positions. Future work should explore more robust camera-based perception for in-the-wild deployment. Meanwhile, the observed trade-off between responsiveness and style fidelity suggests the need for more effective learning algorithms that can better balance real-time adaptability with motion consistency. Additionally, developing a motion capture-free localization pipeline for robust ball and robot state estimation remains an important direction for real-world deployment. On the hardware side, more capable wrist actuation may further enable more stable dexterous serving motions and subsequently improve strike precision. We leave more discussion in Appendix~\ref{appendix:limitation}.

%% file: sec/appendix.tex
\newpage
\begin{center}
    {\LARGE \bfseries Appendix of AdaPT}
\end{center}
\appendix\label{appendix}

\section{Implementation Details}
\subsection{Motion VAE and Residual Tracker }

\paragraph{Motion VAE.} We follow the MotionVAE implementation in Vid2Player3D~\cite{zhang2023learning}, which performs autoregressive motion prediction by conditioning on a short history and predicting the next future frame. Our MotionVAE state design largely follows the original formulation, with two modifications. First, we replace joint rotations with joint angles. Second, we remove global position information since reliable estimation of absolute court coordinates is not available in our pipeline. However, we find our current representation is sufficient for learning effective rally skills. To mitigate the loss of global orientation, we apply data augmentation by randomly rotating the initial frame of each motion clip along the yaw axis within $[-15^\circ, 15^\circ]$, improving coverage over different court-facing directions.

In addition, we extend the MotionVAE prediction targets with four auxiliary dimensions to better structure the latent motion space and facilitate reward design. Specifically, we predict a 2D phase variable normalized to $[0, 2\pi]$, where $\pi$ corresponds to the ball-strike frame, as well as two categorical attributes: spin type and stroke type. These auxiliary predictions provide additional supervision signals and enable easy reward design for rally.

\paragraphbegin{Residual Tracker.} Compared to the Motion Tracker in the first stage, the Residual Tracker additionally incorporates the ball position and the target landing point as observations. It outputs a 27-dimensional residual dof-position within the range of \([-1, 1]\). To encourage the residual action to be as small as possible, an L2 loss that penalizes deviations from zero is added as an optimization objective. Additionally, the Residual Tracker also serves as a planner by outputting an extra phase increment \(\alpha_t\). This phase increment is only active during the interval $[\phi_{\text{release}}, \phi_{\text{hit}} + \Delta\phi_{\text{delay}}]$, where $\phi_{\text{release}}$ denotes the ball release phase, $\phi_{\text{hit}}$ is the striking phase, and $\Delta\phi_{\text{delay}}$ is a constant phase offset. During the remaining stages, we use the fixed phase increment.

\subsection{Reward Functions.}
\paragraph{Rally.}
The rally reward encourages solid racket--ball contact, deep shots after impact, and accurate opponent-court landing. The reward terms are listed below (\cref{tab:rally-reward}). 

\begin{itemize}[leftmargin=4mm,itemsep=1pt]
    \item \textbf{Racket--ball rewards} $R_{\mathrm{racket}}$ ($\omega_{\mathrm{racket}}=5.0$) give dense guidance before a valid strike and a constant reward afterward. Before contact, the agent is encouraged to bring the racket close to the ball and to align with the MVAE swing phase, with forehand/backhand-specific phase targets selected from the predicted skill label. A segment-persistent hit latch is triggered when geometric and kinematic contact criteria are met (distance, racket-face orientation, and approach speed, or custom racket--ball collision when enabled). Once latched, the term stays active until the next incoming-ball segment.

    \item \textbf{Landing rewards} $R_{\mathrm{landing}}$ ($\omega_{\mathrm{landing}}=100.0$) supervise opponent-court placement using a single ball-dynamics rollout per segment. Shortly after impact, the current ball state is rolled forward under the same aerodynamic and bounce model as in the simulation to estimate bounce location and trajectory apex. The target landing point is defined relative to the robot's base on the opponent's half of the court (depth and lateral centering). Landing quality combines position error and apex-height shaping. A \textbf{net-crossing constraint} requires the predicted trajectory to clear the net by at least $1.1\,\mathrm{m}$. Otherwise, the landing reward is zeroed. The resulting score is cached and applied densely for the remainder of the segment.

    \item \textbf{Forward-velocity reward} $R_{\mathrm{forward}}$ ($\omega_{\mathrm{forward}}=50.0$) activates only after a valid hit and encourages deep shots along the forward court direction.
\end{itemize}

\begin{table*}[h]
    \centering
    \caption{Reward functions for rally training.}
    \vspace{-0.1in}
    \setlength{\tabcolsep}{25pt}
    \renewcommand{\arraystretch}{1.1}
    \resizebox{1\linewidth}{!}{
    \begin{tabular}{l l l}
    \toprule
    Term & Weight & Description \\
    \midrule
    Racket--ball & $5.0$ & Racket--ball contact and MVAE phase alignment; latched after a valid hit. \\
    Opponent landing & $100.0$ & Predicted opponent-court bounce; must clear the net ($\ge 1.1\,\mathrm{m}$). \\
    Forward velocity & $50.0$ & Deep forward ball speed after hit. \\
    Spin selection & $-1.0$ & MVAE spin-skill prediction should be consistent with the desired one. \\
    Self-collision & $-10.0$ & Robot self-collisions. \\
    Facing backward & $-20.0$ & Base facing away from opponent court. \\
    \bottomrule
    \end{tabular}}
    \label{tab:rally-reward}
    \vspace{-0.1in}
\end{table*}

\paragraphbegin{Serve.}
The rewards for serving training consists of four components that jointly encourage accurate motion tracking, stable ball tossing, successful ball striking, and controlled ball landing (\cref{tab:serving-reward}). 

\begin{itemize}[leftmargin=4mm,itemsep=1pt]
    \item \textbf{Tracking rewards} follow BeyondMimic~\cite{Liao2025BeyondMimicFM} to preserve reference motion. Since the wrist motion plays a key role in professional serving style, we additionally introduce a \textbf{wrist tracking reward} $R_{\mathrm{wrist}}$ ($\omega_{\mathrm{wrist}}=1.0$) on the striking wrist DoF positions.

    \item \textbf{Tossing rewards} $R_{\mathrm{toss}}$ ($\omega_{\mathrm{toss}}=5.0$) enforce accurate ball release position and velocity tracking. The target toss trajectory is constructed from the dataset by extracting the palm position at release and the racket position at impact. Assuming ballistic motion under gravity, we compute the reference trajectory using a parabolic motion model. To improve robustness under sim-to-real gap in ball handling, we further introduce a \textbf{palm-upward reward} $R_{\mathrm{palm\text{-}up}}$ ($\omega_{\mathrm{palm\text{-}up}}=1.5$), which encourages an upward palm orientation before the toss.

    \item \textbf{Hitting reward} $R_{\mathrm{hit}}$ ($\omega_{\mathrm{hit}}=3.0$) is a binary reward indicating whether the racket successfully strikes the ball. Once triggered, it remains active until episode termination. A hit is considered valid if: (1) the distance between ball and racket center is below $d_{\text{threshold}}=0.15\,\mathrm{m}$, (2) the timestep lies within a $t_h=0.3\,\mathrm{s}$ window around the estimated hit time, and (3) the pre-impact ball apex exceeds $h_{\text{threshold}}=1.8\,\mathrm{m}$.

    \item \textbf{Bounce reward} $R_{\mathrm{bounce}}$ ($\omega_{\mathrm{bounce}}=6.0$) penalizes the error between predicted and target landing positions. To enable efficient supervision, we estimate the landing position using an analytic ball dynamics model immediately after impact. In addition, a \textbf{net-crossing reward} $R_{\mathrm{net}}$ ($\omega_{\mathrm{net}}=1.0$) is introduced, which encourages the ball to pass over the net with height exceeding 1.1 m.
\end{itemize}

\begin{table*}[h]
    \centering
    \caption{Reward functions for serving training.}
    \vspace{-0.1in}
    \setlength{\tabcolsep}{25pt}
    \renewcommand{\arraystretch}{1.1}
    \resizebox{1\linewidth}{!}{
    \begin{tabular}{l l l}
    \toprule
    Term & Weight & Description \\
    \midrule
    Wrist tracking & $1.0$ &  Tracks striking wrist DoF positions. \\
    Tossing      & $5.0$ & Enforces accurate ball release position and velocity. \\
    Palm-upward  & $1.5$ & Encourages an upward palm orientation before the toss. \\
    Hitting        & $3.0$ & Binary reward for racket-ball strike. \\
    Bounce     & $6.0$ & Penalizes error between predicted and target landing positions. \\
    Net-crossing     & $1.0$ & Encourages the ball to pass over the net with height exceeding $1.1\,\mathrm{m}$. \\
    Action rate & $-0.2$ & Penalizes large action variations between consecutive timesteps. \\
    Joint limits & $-10.0$ & Penalizes joints that exceed their joint limits. \\
    Collisions & $-10.0$ & Penalizes self-collision of the robot and collision between the racket and the ground. \\
    \bottomrule
    \end{tabular}}
    \label{tab:serving-reward}
\end{table*}

\subsection{Baseline Implementation}
\paragraphbegin{Rally.}
We evaluate our method against a set of representative baselines for the rally task, covering motion-free reinforcement learning, adversarial motion priors, large-scale motion prior models, and decoupled planner–tracker systems. Overall, these baselines are selected to assess the roles of motion priors, explicit motion decomposition, and adaptive control in long-horizon interactive rally behaviors. For a fair comparison, all methods share the same  observations, rewards, and low-level action space.

\begin{itemize}[leftmargin=4mm,itemsep=1pt]

    \item \textbf{RL-Scratch}~\cite{schulman2017proximal} learns the policy without any motion data, relying solely on task rewards.

    \item \textbf{AMP}~\cite{peng2021amp} incorporates adversarial motion priors learned from reference motion, using a discriminator over short-horizon 10-step DoF trajectories to shape the policy behavior.

    \item \textbf{PULSE}~\cite{luo2024universal} and \textbf{NCP}~\cite{zhu2023neural} are motion-prior methods with continuous and discrete latent representations, respectively. We follow their official implementation.

    \item \textbf{Vid2Player3D}~\cite{zhang2023learning} is a simulation-based framework for humanoid tennis that employs a decoupled planner–tracker pipeline without adaptive mechanisms.
\end{itemize}

\paragraphbegin{Serve.}
We compare our method against representative baselines that learn with or without a single reference motion. These span three categories: 
(1) task-driven pure RL, 
(2) motion priors as rewards, and 
(3) motion-tracking approaches. 
These baselines are selected to evaluate the contributions of our proposed two-stage training and adaptive mechanism.

\begin{itemize}[leftmargin=4mm,itemsep=1pt]
    \item \textbf{RL-Scratch}~\cite{schulman2017proximal} removes all tracking-related rewards and learns the serving skill purely from task rewards. Specifically, only the ball-hitting reward and the landing-point reward are retained.

    \item \textbf{AMP}~\cite{peng2021amp} employs an adversarial motion prior with a discriminator conditioned on 10-step DoF position trajectories. The AMP reward and task reward are weighted by 0.1 and 0.9, respectively.

    \item \textbf{AdaMimic}~\cite{huang2025towards} follows a two-stage training procedure. In stage~1, a pure motion-tracking controller is trained without task rewards or speed adaptation, whereas \ours learns speed-adaptive tracking already in this stage. In stage~2, the controller is conditioned on the predicted high-level adaptation variable $\alpha_t$ to modulate speed throughout the serving motion.

    \item \textbf{DeepMimic}~\cite{peng2018deepmimic} adopts a single-stage training scheme that jointly optimizes tracking and task rewards without a planner. \textbf{DeepMimic-Distill} further distills a student policy from the stage~1 tracking policy by using the teacher's actions as an imitation reward during training.
\end{itemize}

\subsection{Domain Randomization.}
\begin{table*}[t]
    \centering
    \caption{Domain randomization sampling distributions for rally training.}
    \label{tab:dr-rally}
    \renewcommand{\arraystretch}{1.0}
    \setlength{\tabcolsep}{15pt}
    \resizebox{0.8\linewidth}{!}{%
    \begin{tabular}{@{}l l@{}}
        \toprule
        \textbf{Domain Randomization} & \textbf{Sampling Distribution} \\
        \midrule
        \multicolumn{2}{@{}l}{\textit{Physical parameters}} \\
        \addlinespace[2pt]
        Foot tangential friction $\mu$
            & $\mathcal{U}[0.3,\,1.2]$ \\
        Encoder bias $\Delta q$
            & $\mathcal{U}[-0.01,\,0.01]$ \\
        Base COM offset $(\Delta x,\,\Delta y,\,\Delta z)$
            & $\mathcal{U}[-0.08,\,0.08]$ \\
        PD gain scale $(k_p,\,k_d)$
            & $\mathcal{U}[0.95,\,1.05]$ \\
        Torque limit scale
            & $\mathcal{U}[0.9,\,1.0]$ \\
        Actuator delay $\ell$ (physics steps)
            & $\mathcal{U}\bigl[0,\,4\cdot\text{decimation}\bigr]$ \\
        Push interval $\Delta t$
            & $\mathcal{U}[1,\,3]\,\mathrm{s}$ \\
        Push root linear velocity $(v_x,\,v_y,\,v_z)$
            & $\mathcal{U}[-0.5,\,0.5],\;
               \mathcal{U}[-0.5,\,0.5],\;
               \mathcal{U}[-0.2,\,0.2]$ \\
        Push root angular velocity $(\omega_{\mathrm{roll}},\,\omega_{\mathrm{pitch}},\,\omega_{\mathrm{yaw}})$
            & $\mathcal{U}[-0.5,\,0.5],\;
               \mathcal{U}[-0.5,\,0.5],\;
               \mathcal{U}[-0.8,\,0.8]$ \\
        Joint position observation noise
            & $\mathcal{U}[\pm 0.01]$ \\
        Joint velocity observation noise
            & $\mathcal{U}[\pm 0.5]$ \\
        Projected gravity observation noise
            & $\mathcal{U}[\pm 0.05]$ \\
        Root quaternion observation noise
            & $\mathcal{U}[\pm 0.03]$ \\
        Base angular velocity observation noise
            & $\mathcal{U}[\pm 0.2]$ \\
        MVAE init root offset $(\Delta x,\,\Delta y)$
            & $\mathcal{U}[-0.5,\,0.5]\,\mathrm{m}$ \\
        MVAE init yaw offset $\Delta\psi$
            & $\mathcal{U}[-30^\circ,\,30^\circ]$ \\
        \midrule
        \multicolumn{2}{@{}l}{\textit{Ball trajectory \& dynamics}} \\
        \addlinespace[2pt]
        Serve spawn $(x_{\mathrm{rel}},\,y_{\mathrm{rel}},\,z_{\mathrm{rel}})$
            & $\mathcal{U}[6,\,15],\;
               \mathcal{U}[-2,\,2],\;
               \mathcal{U}[0.6,\,1.7]\,\mathrm{m}$ \\
        Flight time $T$
            & $\mathcal{U}[0.7,\,1.4]\,\mathrm{s}$ \\
        Landing target $(x_{\mathrm{rel}},\,y_{\mathrm{rel}},\,z_{\mathrm{tgt}})$
            & $\mathcal{U}[-1,\,3],\;
               \mathcal{U}[-1.5,\,1.5],\;
               0\,\mathrm{m}$ \\
        Re-serve interval
            & $\mathcal{U}[2.0,\,2.5]\,\mathrm{s}$ \\
        Drag coefficient $C_d$
            & $\mathcal{U}[0.4,\,0.8]$ \\
        Ground restitution $e$
            & $\mathcal{U}[0.5,\,0.75]$ \\
        Horizontal bounce damping $d_{xy}$
            & $\mathcal{U}[0.6,\,0.8]$ \\
        Post-bounce velocity noise $(\Delta v_x,\,\Delta v_y,\,\Delta v_z)$
            & $\mathcal{U}[-0.25,\,0.25]\,\mathrm{m/s}$ \\
        Ball position noise $\Delta\mathbf{p}$
            & $\mathcal{U}[-0.02,\,0.02]^3$ \\
        Ball velocity scale
            & $\mathcal{U}[0.9,\,1.1]$ \\
        Ball packet drop
            & 
               $L\sim\mathcal{U}[3,\,5]$ steps ;\; $p_{\mathrm{trigger}}=0.1$; \\
        \bottomrule
    \end{tabular}%
    }
\end{table*}

\begin{table*}[t]
    \centering
    \caption{Domain randomization sampling distributions for serve training.}
    \vspace{-0.1in}
    \label{tab:dr-serve}
    \renewcommand{\arraystretch}{1.0}
    \setlength{\tabcolsep}{15pt}
    \resizebox{0.85\linewidth}{!}{%
    \begin{tabular}{@{}l l@{}}
        \toprule
        \textbf{Domain Randomization} & \textbf{Sampling Distribution} \\
        \midrule
        \multicolumn{2}{@{}l}{\textit{Physical parameters}} \\
        \addlinespace[2pt]
        Foot tangential friction $\mu$
            & $\mathcal{U}[0.3,\,1.2]$ \\
        Encoder bias $\Delta q$
            & $\mathcal{U}[-0.01,\,0.01]$ \\
        Base COM offset $(\Delta x,\,\Delta y,\,\Delta z)$
            & $\mathcal{U}[-0.08,\,0.08]$ \\
        PD gain scale $(k_p,\,k_d)$
            & $\mathcal{U}[0.95,\,1.05]$ \\
        Torque limit scale
            & $\mathcal{U}[0.9,\,1.0]$ \\
        Actuator delay $\ell$ (physics steps)
            & $\mathcal{U}\bigl[0,\,2\cdot\text{decimation}\bigr]$ \\
        Push interval $\Delta t$
            & $\mathcal{U}[1,\,3]\,\mathrm{s}$ \\
        Push root linear velocity $(v_x,\,v_y,\,v_z)$
            & $\mathcal{U}[-0.5,\,0.5],\;
               \mathcal{U}[-0.5,\,0.5],\;
               \mathcal{U}[-0.2,\,0.2]$ \\
        Push root angular velocity $(\omega_{\mathrm{roll}},\,\omega_{\mathrm{pitch}},\,\omega_{\mathrm{yaw}})$
            & $\mathcal{U}[-0.52,\,0.52],\;
               \mathcal{U}[-0.52,\,0.52],\;
               \mathcal{U}[-0.78,\,0.78]$ \\
        Joint position observation noise
            & $\mathcal{U}[\pm 0.01]$ \\
        Joint velocity observation noise
            & $\mathcal{U}[\pm 0.5]$ \\
        Projected gravity observation noise
            & $\mathcal{U}[\pm 0.05]$ \\
        Base angular velocity observation noise
            & $\mathcal{U}[\pm 0.2]$ \\
        \midrule
        \multicolumn{2}{@{}l}{\textit{Motion reference initialization}} \\
        \addlinespace[2pt]
        Root position offset $(\Delta x,\,\Delta y,\,\Delta z)$
            & $\mathcal{U}[-0.05,\,0.05],\;
               \mathcal{U}[-0.05,\,0.05],\;
               \mathcal{U}[-0.01,\,0.01]$ \\
        Root orientation offset $(\Delta\mathrm{roll},\,\Delta\mathrm{pitch},\,\Delta\mathrm{yaw})$
            & $\mathcal{U}[-0.1,\,0.1],\;
               \mathcal{U}[-0.1,\,0.1],\;
               \mathcal{U}[-0.2,\,0.2]$ \\
        Root linear velocity offset $(\Delta v_x,\,\Delta v_y,\,\Delta v_z)$
            & $\mathcal{U}[-0.5,\,0.5],\;
               \mathcal{U}[-0.5,\,0.5],\;
               \mathcal{U}[-0.2,\,0.2]$ \\
        Root angular velocity offset $(\Delta\omega_{\mathrm{roll}},\,\Delta\omega_{\mathrm{pitch}},\,\Delta\omega_{\mathrm{yaw}})$
            & $\mathcal{U}[-0.52,\,0.52],\;
               \mathcal{U}[-0.52,\,0.52],\;
               \mathcal{U}[-0.78,\,0.78]$ \\
        Joint position offset
            & $\mathcal{U}[-0.1,\,0.1]$ \\
        \midrule
        \multicolumn{2}{@{}l}{\textit{Ball trajectory \& observations}} \\
        \addlinespace[2pt]
        Toss time delay $\Delta t_{\mathrm{toss}}$
            & $\mathcal{U}[-0.02,\,0.05]\,\mathrm{s}$ \\
        Toss velocity perturbation $(\Delta v_x,\,\Delta v_y,\,\Delta v_z)$
            & $\mathcal{U}[-0.1,\,0.1]$ \\
        Landing target $(x_{\mathrm{tgt}},\,y_{\mathrm{tgt}})$
            & $\mathcal{U}[6.5,\,20.0],\;
               \mathcal{U}[-1.0,\,1.0]\,\mathrm{m}$ \\
        Ball position observation noise
            & $\mathcal{U}[\pm 0.1]$ \\
        Ball position observation delay $\ell$ (physics steps)
            & $\mathcal{U}[0,\,5]$ \\
        Ball observation dropout
            & $p_{\mathrm{drop}}=0.1$ \\
        \bottomrule
    \end{tabular}%
    }
    \vspace{-0.1in}
\end{table*}

We present the domain randomization terms for rally and serve tasks, including physical parameters, ball trajectories and motion reference initialization, in \cref{tab:dr-rally} and \cref{tab:dr-serve}, respectively. Importantly, we find that adding noise to the ball trajectory is essential for robust learning. Even in motion-capture settings, the ball observations are subject to noise, packet loss, and latency, which can significantly degrade policy performance if not accounted for during training.  In addition, due to the limited vertical field of view in our setup, we mask ball observations when the ball exceeds a height threshold. These engineering efforts are crucial in practice and have a significant impact on the final performance. The performance of the rally will drop from 80\% to 55\% without it.

\subsection{PPO Implementation.} 
Our PPO implementation follows the framework outlined in~\cite{Rudin2021LearningTW}. The actor and critic networks consist of 3-layer MLPs. Each iteration includes 25 steps per environment, with 5 learning epochs and 4 mini-batches per epoch. The discount factor $\gamma$ is set to 0.99, and the clip ratio is set to 0.2. In the rally task, the entropy coefficient is set as 0 for training stability, following~\cite{zhang2023learning}, and 0.01 in other trainings by default. In the serving task, the weight of the L2 loss for the residual action is 0.01.

\section{Implementation Details in the Real}

\subsection{Ball Localization with Cameras}

\paragraphbegin{Ball detection.}
We also evaluate the motion-based ROI extraction and HSV-based color thresholding used in ESTHER~\cite{zaidi2023athletic}. However, we find this heuristic-based approach more sensitive to visual noise and background motion. Therefore, our primary deployment pipeline uses a YOLO-based detector~\cite{redmon2016you} followed by stereo triangulation for 3D ball localization.

\paragraphbegin{Multi-view 3D reconstruction.}
Given the 2D detections, we estimate 3D ball positions via calibrated multi-view geometry. Each camera provides a candidate observation, which is projected into the world coordinate frame using the pre-computed camera calibration parameters. To improve robustness across viewpoints, we perform confidence-weighted fusion over the six cameras. Specifically, each observation is weighted inversely proportional to the estimated distance between the detected ball and the corresponding camera, ensuring that closer and more reliable views contribute more strongly to the final 3D estimate.

\paragraphbegin{Temporal filtering.}
The fused 3D positions are further processed using an Extended Kalman Filter (EKF) to obtain a temporally consistent ball trajectory. The EKF operates in the world coordinate frame and estimates both position and velocity under a constant-acceleration motion model, which is suitable for airborne ball dynamics under gravity.

\subsection{Ball Future Trajectory Estimation.}
The ball's current position is expressed in a robot-centric frame,
$\mathbf{p}_t = \mathbf{x}^{\mathrm{ball}}_w - \mathbf{x}^{\mathrm{base}}_w$,
which provides local spatial awareness for interaction. Future ball trajectories are obtained by rolling out a physics-based prediction model that approximates ball dynamics under gravity, aerodynamic drag, and ground contact. We omit the rotation of the ball because of the difficulty of the measurement in the real world. The continuous-time dynamics can be written as
\[
\dot{\mathbf{v}} = \mathbf{g} - k C_d \|\mathbf{v}\| \mathbf{v},
\]
where $\mathbf{g}$ is gravity, and $k$ and $C_d$ models quadratic air resistance. Ground interactions are handled using a simple restitution model with velocity reflection and damping. To provide anticipatory information, we append a $K$-step rollout of predicted future ball positions $\{\hat{\mathbf{u}}_{t+1}, \ldots, \hat{\mathbf{u}}_{t+K}\}$ computed from this dynamics model.

To improve robustness to perception and modeling errors, we inject stochastic perturbations into the ball command, including position and velocity noise, short observation delays, and intermittent packet loss, simulating realistic estimator uncertainty. The corresponding noise ranges and domain randomization parameters are summarized in~\cref{tab:dr-rally}. On the real robot, we estimate ball velocity using a five-step temporal window over consecutive position observations, providing a smoothed velocity estimate for the command representation.

\subsection{System Identification}

We collect real-world motion capture data from 20 ball trajectories on the physical court to identify key parameters of the ball dynamics model. Given the analytic ball model used in simulation, we fit two primary physical parameters: the coefficient of restitution and the horizontal damping coefficient. These parameters govern the ball’s post-impact energy loss and in-plane velocity decay, respectively. Parameter estimation is performed by minimizing the trajectory discrepancy between the observed motion capture data and the model-predicted rollout under identical initial conditions. The resulting calibrated parameters are used consistently in simulation with randomization to reduce the sim-to-real gap in ball dynamics.

\section{More Limitations}\label{appendix:limitation}

\begin{wrapfigure}{r}{0.4\linewidth}
    \centering
    \vspace{-0.15in}

    \setlength{\tabcolsep}{7.5pt}
    \resizebox{\linewidth}{!}{
    \begin{tabular}{lcc}
        \toprule
         & Sim. & Real\\
        \midrule
        $E_{\mathrm{bo}} \downarrow$ & 2.81\ci{1.73} & 3.90\ci{2.74}  \\
        \bottomrule
    \end{tabular}
    }

    \vspace{5pt}

    \begin{minipage}[c]{0.48\linewidth}
        \centering
        \includegraphics[width=\linewidth]{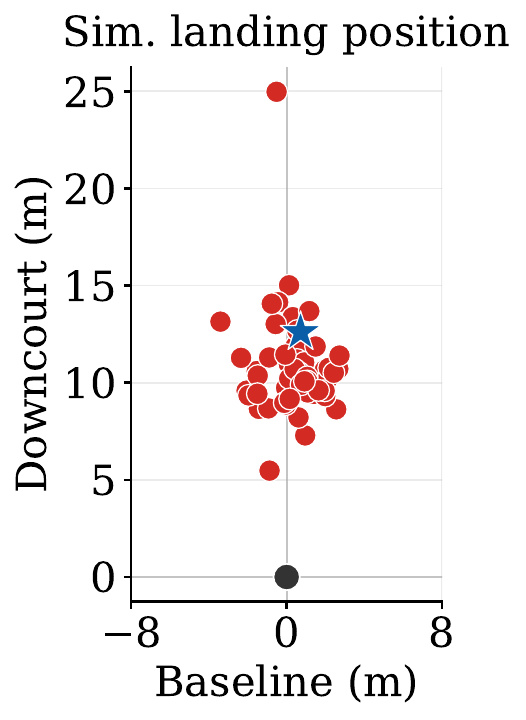}
    \end{minipage}
    \hfill
    \begin{minipage}[c]{0.48\linewidth}
        \centering
        \includegraphics[width=\linewidth]{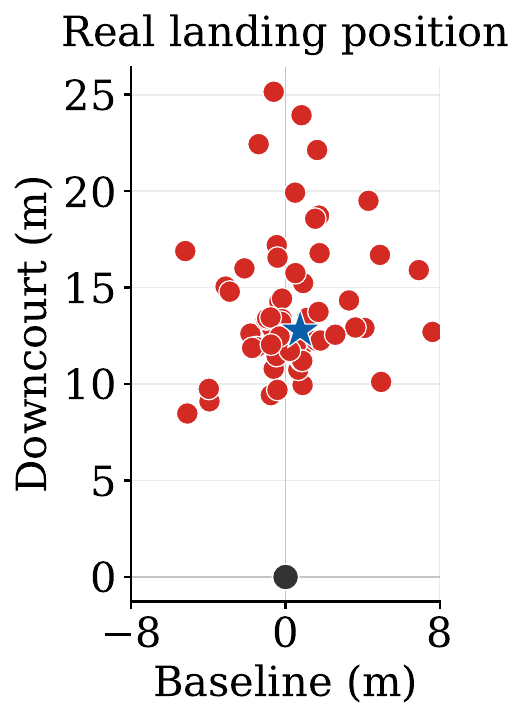}
    \end{minipage}

    \vspace{-0.18in}
\end{wrapfigure}
\textbf{Return quality.} The real-world return
quality is less consistent than in simulation and is sensitive to the incoming ball
landing position and velocity. We  quantitatively evaluate
$E_{\mathrm{bo}}$ (\textcolor{gray}{Tab.~$\rightarrow$}) and visualize the landing
locations in \textcolor{gray}{Figure.~$\rightarrow$}. This raises the difficulty of human-humanoid play, and we will address this problem in the future.

\paragraphbegin{Limited locomotion ability}. We  share another important trade-off: autoregressive generation enables cyclic
\begin{wrapfigure}{r}{0.34\linewidth}
    \centering
    \vspace{-0.2in}
    \includegraphics[width=\linewidth]{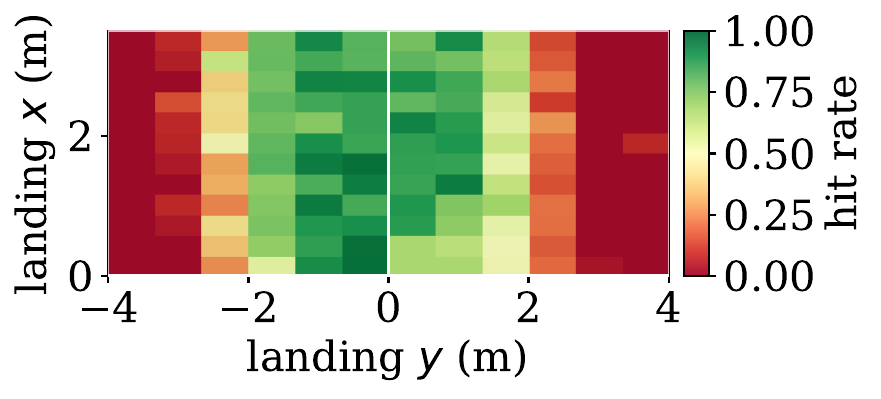}
    \vspace{-0.1in}
    \vspace{-0.25in}
    \label{fig:rlscratch}
\end{wrapfigure}
 rally motions but sacrifices locomotion capability. \textcolor{gray}{Fig. $\rightarrow$} shows lower hit rates of AdaPT at distant positions. This is less noticeable in NCP or PULSE. Addressing this problem is valuable in the future. 
 
\section{More Real-world Experiments}

\begin{figure}[h]
    \centering
    \vspace{-0.15in}
    \includegraphics[width=1\linewidth]{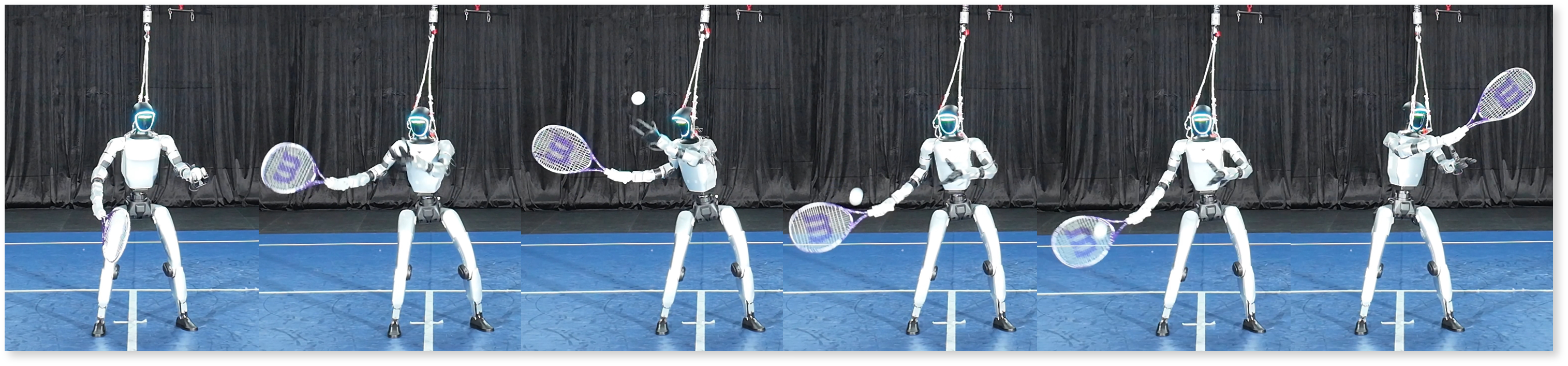}
    \caption{Amateur serving motion using the Unitree Dex-3 end-effector.}
    \label{fig:dex3_serve}
    \vspace{-0.2in}
\end{figure}

\paragraph{Serve with a dexterous hand.} For the serving task, our method supports various end-effectors for the tossing arm. In initial tests, we used Unitree's Dex-3 as the end-effector. However, its substantial weight poses a collision risk during real-robot deployment, as professional players typically swing the tossing arm at high speeds. To mitigate this, we recorded an amateur serving motion with reduced tossing arm swing and tested it on the real robot (\cref{fig:dex3_serve}), achieving a 90\% success rate over 10 trials. In future work, we plan to adopt a lighter dexterous hand to enable more professional serving motions.

\begin{figure}[h]
    \centering
    \vspace{-0.25in}
    \includegraphics[width=0.75\linewidth]{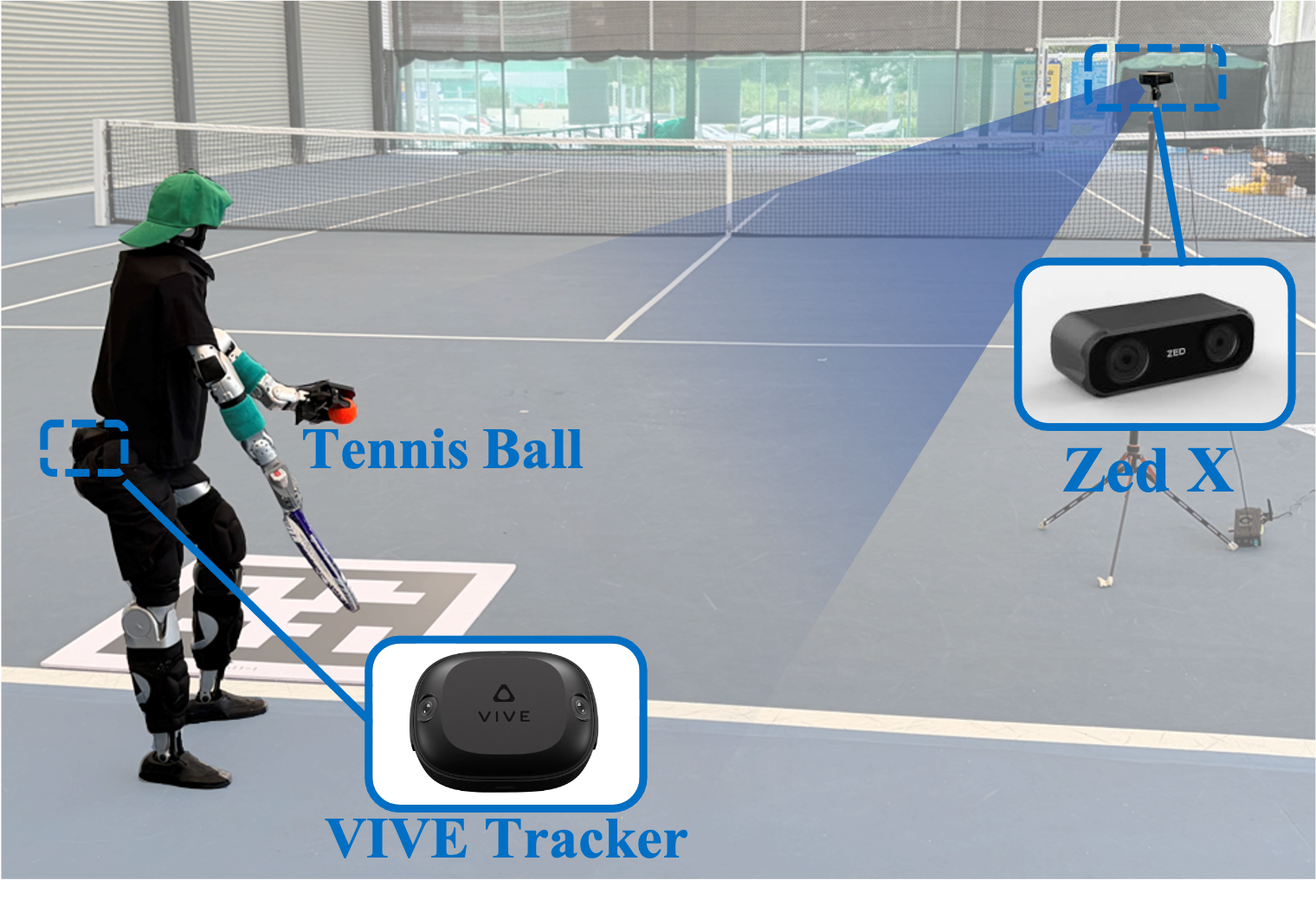}
    \vspace{-0.15in}
    \caption{The setup of in-the-wild serve.}
    \label{fig:inthewild}
\end{figure}

\paragraph{In-the-wild serve.} We visualize the setup in~\cref{fig:inthewild}.


